%% file: main.tex
\documentclass[11pt]{article}
\usepackage[preprint]{acl}
\usepackage{times,latexsym}
\usepackage[T1]{fontenc}
\usepackage[utf8]{inputenc}
\usepackage{microtype,inconsolata}
\usepackage{amsmath,amssymb,booktabs,tabularx,array,graphicx,xcolor}
\usepackage{placeins}
\newcommand{\method}{\textsc{NAC}}
\newcommand{\act}{\textsc{act}}
\newcommand{\wait}{\textsc{wait}}
\newcommand{\spec}{\textsc{spec}}

\newcommand{\czh}[1]{}
\newcommand{\peng}[1]{}
\newcommand{\chenzh}[1]{}
\newcommand{\Z}[1]{}

\title{Neural Agent Compiler: Learning to Schedule Soft Dependencies}

\author{
  Bo Yan\textsuperscript{1},
  Peng Wang\textsuperscript{2},
  Zihan Chen\textsuperscript{2},
  Ziwen Kan\textsuperscript{1},
  Zhen Tan\textsuperscript{3},
  Huiyuan Chen\textsuperscript{4},\\
  \textbf{Bo Sun}\textsuperscript{5},
  \textbf{Shuai Shao}\textsuperscript{5},
  \textbf{Tianlong Chen}\textsuperscript{6},
  \textbf{Song Wang}\textsuperscript{1} \\
  \textsuperscript{1}University of Central Florida \quad
  \textsuperscript{2}University of Virginia \quad
  \textsuperscript{3}Stevens Institute of Technology \\
  \textsuperscript{4}Amazon \quad
  \textsuperscript{5}Meta \quad
  \textsuperscript{6}The University of North Carolina at Chapel Hill \\
  \small\texttt{\{bo.yan, ziwen.kan\}@ucf.edu}, \quad
  \texttt{\{pw7nc, brf3rx\}@virginia.edu}, \\
  \small\texttt{ztan12@stevens.edu}, \quad
  \texttt{huiyuach@amazon.com}, \quad
  \texttt{tianlong@cs.unc.edu}, \\
  \small\texttt{\{bsunmiracle, sshao\}@meta.com}, \quad
  \small\texttt{song.wang@ucf.edu}
}

\begin{document}
\maketitle
\input{sections/00_abstract}
\input{sections/01_introduction}
\input{sections/02_related_work}
\input{sections/03_formulation}
\input{sections/04_method}
\input{sections/05_experiments}
\input{sections/06_results}
\input{sections/07_analysis}

\input{sections/08_conclusion}
\input{sections/09_limitations}
\bibliography{references}
\clearpage
\appendix
\raggedbottom
\input{appendix/teaser_candidates_20261008}
\input{appendix/a_training}

\input{appendix/b_results}

\input{appendix/c_statistics}
\input{appendix/d_extensions}

\input{appendix/e_mechanism_summary}

\input{appendix/front_support_main_candidates}

\end{document}

%% file: sections/00_abstract.tex
\begin{abstract}

Large language model agents accomplish complex tasks by coordinating multiple 
reasoning steps and tool operations, which form dependencies 
among their operations. Existing orchestration methods primarily consider hard 
dependencies that determine when an operation can legally begin, but overlook 
\emph{soft dependencies}, i.e., information that is not required for execution 
but may improve an already executable operation. We identify soft-dependency 
scheduling as an important problem in agent execution. 
To this end, we introduce  Neural Agent 
Compiler (NAC), a learned scheduler that predicts the task-quality and resource 
consequences of three execution modes and selects the highest-utility legal 
mode at a scheduling boundary. NAC represents the executable continuation, 
pending information event, and resource state in a shared memory, and uses a 
 utility function to adapt the consequence predictions to different 
resource preferences. Across AppWorld, WebShop, and $\tau^2$-bench, NAC achieves
276/417
AppWorld task completions, a 74.69 WebShop Task Score, and 59/100 $\tau^2$-bench
successes. Additional studies show that the learned scheduler reduces execution 
work, establishing soft-dependency 
scheduling as a promising solution for orchestrating tool-using agents.

\end{abstract}

%% file: sections/01_introduction.tex
\section{Introduction}
\label{sec:intro}

\begin{figure*}[t]
\centering
\includegraphics[width=.9\textwidth]{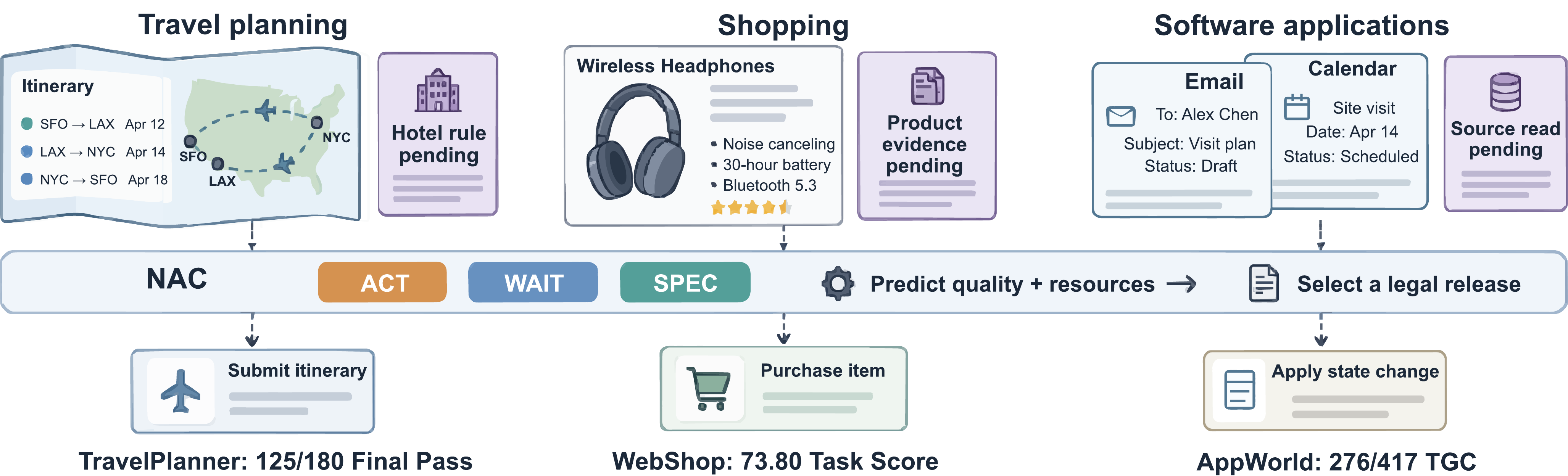}
\caption{\textbf{The same release decision acquires native meaning at each commit boundary.} \method{} selects among three legal ways to incorporate pending evidence into itinerary construction, product selection, and software-application execution; the panels show the corresponding information and commitment boundaries.}
\label{fig:nc2-selected-teaser-c}
\end{figure*}

Large language model (LLM) agents increasingly accomplish complex tasks by coordinating multiple reasoning steps and tool operations \citep{yao2023react,kim2024llmcompiler,zhang2025aflow,hu2025adas}. Such tasks naturally form executable programs, where operations consume intermediate results, produce new information, and depend on one another \citep{kim2024llmcompiler}. Efficient agent execution therefore requires not only determining \emph{what} operations to perform, but also \emph{when} to execute them \citep{xiang2026scheduler,winston2026agentjit,zhang2025planovergraph,kashmira2026contextrender,zhang2025agentprune}. 

Existing dependency-based execution frameworks such as LLMCompiler \citep{kim2024llmcompiler} primarily organize operations according to \emph{hard dependencies}: prerequisites that must be satisfied before an operation can begin. However, satisfying these prerequisites does not necessarily make immediate execution optimal. An operation may already be executable while additional information, still being acquired, could substantially improve its outcome. We refer to such relationships as \emph{soft dependencies}. Unlike hard dependencies, which determine whether an operation can execute, soft dependencies influence whether it 
should execute immediately or defer execution until more information becomes 
available. 

Scheduling under soft dependencies is challenging because the available execution modes trade quality, latency, and revision work. Consider a travel-planning agent that has a route and travel dates but is still waiting for a hotel query that may reveal booking restrictions~\citep{ye2026speculative,hooper2026interaction,yu2026tomasullm,liu2026smc}. It can act immediately with current evidence, wait for the query, or prepare a reversible itinerary while deferring commitment until the missing information arrives~\citep{chen2026aospec,fareed2026costaware,xu2026leap}. None is uniformly best: the right choice depends on how likely the pending information is to change the downstream decision, how much work can be reused, and what each mode costs. Optimizing agent execution therefore requires reasoning about both the \emph{value of future information} and the \emph{cost of acting without it} \citep{howard1966information,kim2026learningask,rajput2026pareto,yilmaz2026lavoir,hu2026routeaudit}.

Our key insight is that this scheduling problem can be addressed by learning the consequences of alternative execution strategies without committing to a fixed execution policy. 
These consequences capture how a scheduling decision affects 
eventual task quality, latency, computation, and revision. The scheduler can then 
compare the expected utility of alternative execution strategies under the current 
state and resource constraints.

Based on this insight, we introduce \emph{Neural Agent Compiler} (\method{}),
a lightweight learned compiler for scheduling soft dependencies in tool-using agents (Fig.~\ref{fig:nc2-selected-teaser-c}).
\method{} is built around three execution modes: \emph{immediate execution} 
(\act{}), which starts the continuation using currently available evidence; 
\emph{evidence-first execution} (\wait{}), which waits for the pending information; 
and \emph{reversible speculation} (\spec{}), which performs bounded preparatory work 
before committing externally visible effects. Given an executable continuation, a 
pending information event, and the available resource state, \method{} constructs a 
typed representation of the execution boundary and predicts the task quality and 
resource consequences of each mode. A calibrated utility function then selects the 
highest-valued legal mode, and the resulting directive is passed to the  
executor. This design separates learned scheduling from task-specific tool execution 
while providing a common decision framework across environments.
Extensive evaluation is conducted on various benchmarks to verify the effectiveness and efficiency of \method{}.
Our contributions are summarized as follows:
\begin{itemize}
    \item \textbf{Soft-dependency-aware agent compilation.} We formulate agent execution scheduling beyond hard prerequisites by introducing soft dependencies, which capture how pending information can improve operations that are already executable.

    \item \textbf{Learned consequence-based execution scheduling.} We propose \method{}, a lightweight agent compiler that predicts the quality and resource consequences of immediate execution, evidence-first execution, and reversible speculation, and translates these predictions into execution decisions.

    \item \textbf{Evaluation of adaptive execution.} We evaluate the proposed framework across multiple agent environments and provide comparisons examining task quality, execution efficiency, preparation costs, and sensitivity to resource preferences.
\end{itemize}

%% file: sections/02_related_work.tex
\section{Related Work}
\label{sec:related}
\paragraph{Compiling and organizing agent workflows.}
ReAct interleaves reasoning and actions, and LLMCompiler schedules tool dependencies \citep{yao2023react,kim2024llmcompiler}. GPTSwarm and AFlow optimize agent structures and workflows, while ADAS and DyLAN explore automated system design and dynamic agent networks \citep{zhuge2024gptswarm,zhang2025aflow,hu2025adas,liu2024dylan}. LLM-as-Scheduler, MIRROR, and SwarmAgentic respectively study workflow routing, reflection, and population-based system construction \citep{xiang2026scheduler,guo2025mirror,zhang2025swarmagentic}. Plan-over-Graph and Agent JIT compilation focus on parallel or latency-aware scheduling, and ContextRender and AgentPrune reduce execution-context or communication overhead \citep{zhang2025planovergraph,winston2026agentjit,kashmira2026contextrender,zhang2025agentprune}. 

\paragraph{Speculative agent execution.}
Speculative Actions and Speculative Interaction Agents overlap action drafting with tool execution \citep{ye2026speculative,hooper2026interaction}. TomasuLLM executes drafted actions in isolated sandboxes and validates their effects before committing them in trajectory order \citep{yu2026tomasullm}. LEAP learns action proposals from target-actor trajectories \citep{xu2026leap}. Speculative Macro Commit, AOSpec, and cost-aware speculative execution extend this line to multi-action commit, action--observation co-speculation, and explicit resource trade-offs \citep{liu2026smc,chen2026aospec,fareed2026costaware}. These works supply close precedents for early execution and validated commitment. \method{} uses terminal quality and resource predictions to choose among immediate execution, waiting, and reversible preparation. Speculation is one available schedule, and the learned comparison estimates when its preparation is useful.

\paragraph{Learning the value of information.}
Expected information value quantifies whether new evidence improves downstream decisions \citep{howard1966information}. Prior work operationalizes this by learning when to inspect external state \citep{di2026budgetpm,fang2026budgetcontrol,zhang2026harnessvalue}, balancing clarification benefits against budget constraints \citep{yilmaz2026lavoir,park2026sag,nguyenhien2026revoir,kim2026learningask,rajput2026pareto}, or optimizing runtime resource accounting \citep{ofengenden2026agenttime,hu2026routeaudit}. Extending this principle, \method{} evaluates the value of waiting directly against execution and revision costs, drawing on incremental computation to preserve reusable intermediate work \citep{hammer2014adapton,mokhov2018build}.


\paragraph{Native task evaluation.}
WebShop~\citep{yao2022webshop} and AppWorld~\citep{trivedi2024appworld} expose constrained outcomes through purchase reward and application-state assertions. GAIA~\citep{mialon2024gaia} and Flex-TravelPlanner~\citep{oh2025flextravelplanner} broaden evaluation to general-assistant and flexible planning settings, while AppWorld-UL~\citep{chen2026appworldul} and $\tau^2$-bench~\citep{barres2025tau2} emphasize diverse user interaction and dual-control conversational environments. 

%% file: sections/03_formulation.tex
\section{Preliminaries}
\label{sec:formulation}



\subsection{Problem Formulation} 

An agent program contains operations connected by the inputs and observations they consume. 
The runtime supplies a ready continuation and a designated information source; 
\method{} chooses when to release that continuation relative to the information. 
This is the compiler's learned scheduling boundary.

Let $s^-=(x,P,L,b,\phi(\omega))$ contain the request $x$, remaining program $P$, 
 observed execution ledger $L$, resource budget $b$, and a public description of
information event $\omega$. Here $\phi(\omega)$ denotes the encoded public event description, which can include a query's arguments and
expected result type. When the event returns observation $o$, the continuation can 
use $s^+=(s^-,o)$. We condition on a scheduling boundary supplied by the native 
interface. It identifies a ready continuation, a designated pending information event, 
the available resource state, and the legal execution modes. If several sources 
contribute to the event, their public description may be aggregated in $\phi(\omega)$; 
the current controller does not independently select or order those sources, discover 
the full dependency graph, or jointly schedule arbitrary executable operations.

For each legal execution mode $a\in\mathcal A$ (ACT, WAIT, or SPEC), define the consequence vector
\begin{equation}
 y_a=(q_a,c_{a1},c_{a2},c_{a3},c_{a4}).
 \label{eq:consequences}
\end{equation}
Here $q_a$ measures native task quality; the remaining coordinates describe delay, 
work, or revision consequences in the interface's units. Observation masks identify 
the coordinates measured in each task family. Scheduling uses a linear utility,
\begin{equation}
 U_a=\alpha q_a-\sum_{j=1}^4\lambda_j c_{aj},
 \quad \alpha>0,\quad\lambda_j\geq0.
 \label{eq:utility}
\end{equation}
The value of waiting is 
$\Delta_{\wait}(s^-)=\mathbb E[U_{\wait}-U_{\act}\mid s^-]$. 
It prices the information through the quality and work of the continuation it changes, 
including acquisition and delay. The speculative mode supplies a third comparison when 
early preparation remains useful after observation.

\subsection{Execution Setting}

Hard dependencies establish the prerequisites for a legal operation. Soft dependencies 
connect an executable operation to additional evidence that can change its continuation. 
A soft dependency is defined relative to this downstream use: the event may improve the 
selected work even though its initial operation is ready. The execution-mode set is 
$\mathcal A=\{\act,\wait,\spec\}$. $\act{}$ starts the continuation from $s^-$ and 
permits overlap with information acquisition where the native interface supports it. 
$\wait{}$ releases the continuation after $o$ becomes available, serializing it behind 
that information. $\spec{}$ starts a bounded reversible prefix and holds its externally 
visible effect until the executor incorporates $o$. The native interface defines the 
prefix, legal operations, and commit condition. Hard prerequisites constrain each 
operation within the selected continuation.

The executor supplies a legality mask $m(a)$. The neural core predicts all three 
consequence vectors from the same $s^-$ and compiles them into the directive
\begin{equation}
 a^*=\arg\max_{a:\,m(a)=1}\left(\widehat U_a-\kappa_a\right),
 \label{eq:selection}
\end{equation}
where $\widehat U_a$ is the utility computed from the predicted consequence vector $\widehat y_a$, and $\kappa_a$ is a fixed calibration offset. A positive offset for an early mode
requires a larger predicted gain before departing from waiting. Utility prices express 
the desired quality--resource trade-off, and calibration sets the evidence required to 
change the reference schedule. The measured systems apply this learned decision once 
per task; the native executor carries out the resulting continuation.

%% file: sections/04_method.tex
\section{Neural Agent Compiler}
\label{sec:method}

\begin{figure*}[t]
\centering
\includegraphics[width=.95\textwidth]{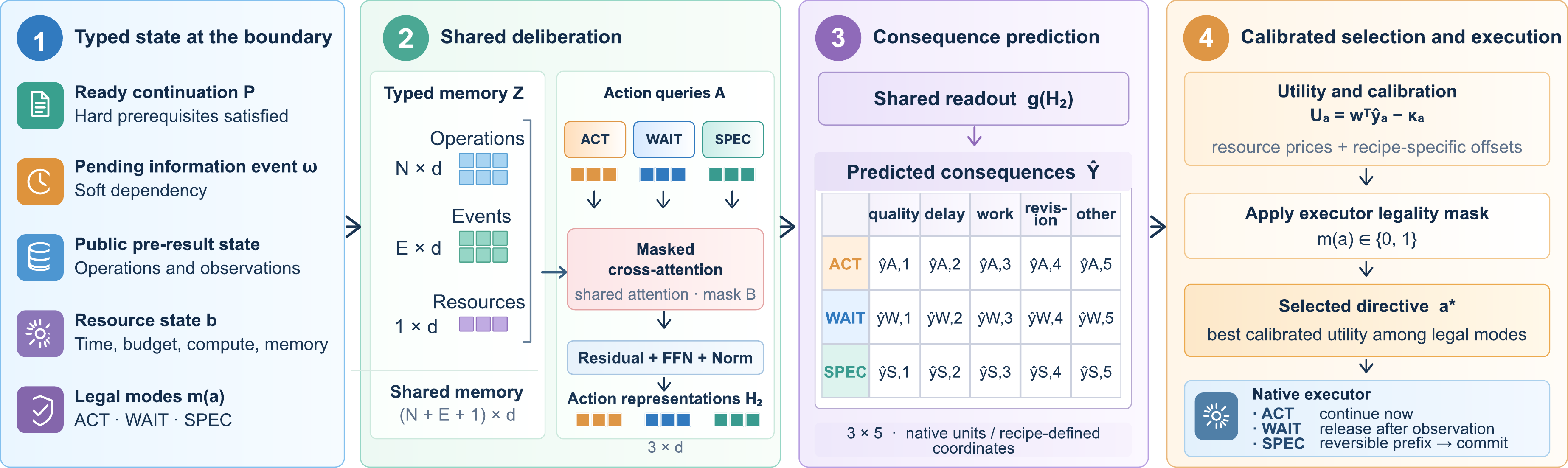}
\caption{\textbf{Neural Agent Compiler at one scheduling boundary.} The native interface supplies the ready continuation, designated information event, public state, and legal modes. Typed encoders form a shared memory; three action queries read it through a deliberation block; a shared readout predicts quality and resource consequences. Calibrated utility and legality produce the directive that the native executor applies.}
\label{fig:controller}
\end{figure*}

\paragraph{Overview.}
\method{} compiles one native scheduling boundary into an execution directive. The native interface supplies a ready continuation, a designated pending information event, the public pre-result state, and the legal execution modes. The controller reads this state once, predicts the consequences of \act{}, \wait{}, and \spec{}, and passes the selected mode to the native executor. The controller therefore ends at action selection: it does not acquire information, execute tools, or decide the full agent program.
\begin{equation}
\begin{aligned}
s^- &\longrightarrow Z \longrightarrow H_2 \longrightarrow \widehat Y,\\
a^* &\in\{\act,\wait,\spec\}.
\end{aligned}
\label{eq:method-overview}
\end{equation}
Here $Z$ is the typed memory, $H_2$ is the final action-query representation, and $\widehat Y$ is the matrix of predicted consequence vectors. The four stages are typed memory, action-conditioned deliberation, consequence prediction, and calibrated release. Figure~\ref{fig:controller} summarizes this shared computation; Appendix~\ref{app:training} gives the measured parameterizations, target maps, and native recipes.

\subsection{Typed state and action-conditioned deliberation}
The input interface supplies operation rows $X_N\in\mathbb R^{N\times d_N}$, information-event rows $X_E\in\mathbb R^{E\times d_E}$, a resource row $x_R\in\mathbb R^{d_R}$, and validity masks. Operation rows describe available work and observed items; event rows describe the pending information boundary; the resource row describes constraints and consumed budget. At selection time the memory contains only public pre-result information.
\begin{equation}
\begin{split}
 Z=[&E_N(X_N)+e_N;\ E_E(X_E)+e_E;\\
    &E_R(x_R)+e_R]\in\mathbb R^{(N+E+1)\times d}.
\end{split}
\label{eq:memory}
\end{equation}
Each type encoder $E_k$ maps its rows to width $d$, and the learned type vector $e_k$ identifies its role. Validity masks remove padded rows while preserving the variable-size native inputs. This typed memory lets the same controller relate executable work, pending information, and resource state without giving the returned observation to the predictor.

Let $A\in\mathbb R^{3\times d}$ contain three learned queries, one for each execution mode. They read the same memory and share the attention parameters while retaining their action identities, so the same pending event can support different predictions for acting, waiting, and reversible preparation. The controller uses one action-conditioned block:
\begin{equation}
\begin{aligned}
Q^{(0)}&=A,\\
Q^{(1)}&=\mathcal D_{\theta}\bigl(Q^{(0)},Z;B\bigr).
\end{aligned}
\label{eq:attention}
\end{equation}
Here $B$ is the validity mask. The block uses masked cross-attention to retrieve relevant memory rows, then combines each retrieved representation with its own action query through a feed-forward residual update and normalization. The action queries do not perform a separate self-attention step; their shared memory and shared parameters provide the common comparison basis while their learned identities preserve the three mode-specific views. The detailed projections and benchmark-specific parameterizations are given in Appendix~\ref{app:layers}. The final action representations are
\begin{equation}
H_2=Q^{(1)}.
\label{eq:residual}
\end{equation}
This single forward pass returns one representation for each legal execution mode.

\subsection{Consequence prediction and supervision}
A shared readout maps every final action representation to the same five-coordinate consequence interface:
\begin{equation}
 \widehat Y_a=g(H_{2,a})\in\mathbb R^5,
 \qquad \widehat Y\in\mathbb R^{3\times5}.
 \label{eq:prediction}
\end{equation}
The first coordinate represents native task quality and the remaining coordinates represent delay, work, or revision consequences in the native units. The coordinates are a common output contract rather than five universal physical quantities: target maps, units, normalization, and observation masks are supplied by each native recipe. The controller predicts how a pending observation changes each execution mode, not the content of that observation itself. Decoding the predicted units makes the same consequence readout compatible with different resource prices in Eq.~\eqref{eq:utility}.

Training pairs one public pre-result state with the outcomes of its legal matched continuations. Let $M_{iaj}$ indicate that coordinate $j$ is observed for state $i$ and action $a$. The shared consequence objective is
\begin{equation}
 \mathcal L_{\rm con}=\sum_{j=1}^5
 \frac{\sum_{i,a}M_{iaj}\,\operatorname{Huber}_\delta
 (\widehat y_{iaj}-y_{iaj})}
 {\max(1,\sum_{i,a}M_{iaj})}.
 \label{eq:loss}
\end{equation}
Missing coordinates do not contribute to the loss. Recipe-specific utility or action-supervision terms attach to this common output when their native targets are available; their coefficients, target construction, and optimization schedules are reported in Appendix~\ref{app:training}. The generative actor remains fixed while the controller is fitted, so the comparison isolates the release decision across matched continuations.

\subsection{Calibrated release and native execution}
At deployment, the controller decodes the predicted consequences, applies the utility and legality rules in Eq.~\eqref{eq:selection}, and returns one execution directive. TravelPlanner and AppWorld use zero release offsets. WebShop uses a held-out calibration set to select a \wait{} reference margin, retaining \wait{} unless the predicted departure advantage exceeds that margin or \wait{} is illegal. These rules define the operating point without changing the learned consequence representation.

The native executor then implements the selected mode: \act{} starts the continuation with current evidence, \wait{} releases it after the designated information arrives, and \spec{} performs a bounded reversible prefix before evidence-conditioned commitment. The executor owns information acquisition, hard prerequisites, budgets, permissions, prefix legality, and commitment; the controller supplies only the mode. Benchmark-specific input schemas, target units, branch recipes, and native scoring are given in Appendix~\ref{app:training}.

%% file: sections/05_experiments.tex
\section{Experiments}
\label{sec:experiments}
\subsection{Experimental Settings}

\textbf{1. Datasets and outcomes.} We evaluate \method{} on AppWorld, WebShop, and $\tau^2$-bench, which together provide
1,017 native task IDs across three primary populations. The evaluation uses all 417 
challenge tasks in AppWorld, all 500 official test tasks in WebShop, and a disjoint 
100-task test population in $\tau^2$-bench. AppWorld reports Task Goal Completion (TGC) 
and Scenario Goal Completion (SGC), where each of its 139 scenarios contains three 
tasks. WebShop reports the native purchase reward as a Task Score on a 0--100 scale. 
$\tau^2$-bench reports native dialogue success over 20 airline, 40 retail, and 40 
telecom tasks. Each task contributes its final native outcome, including failed 
completions. \textbf{2. Baselines and comparison protocol.} Each main result compares \method{} with external methods evaluated under their
reported execution settings. AppWorld includes the released reactive, 
plan-and-execute, reflective, and function-call agents \citep{trivedi2024appworld}, 
as well as ReAct \citep{yao2023react}, CodeAct \citep{wang2024codeact}, and LLMCompiler 
\citep{kim2024llmcompiler}. WebShop includes the released rule, imitation, and 
policy-gradient policies \citep{yao2022webshop}, together with LATS 
\citep{zhou2024lats}. $\tau^2$-bench includes native LLMAgent and ReAct evaluated on the 
same 100 test IDs \citep{barres2025tau2,yao2023react}. Tables~\ref{tab:systems}, 
\ref{tab:shopping}, and~\ref{tab:tau2-main} report the complete native outcomes. 
Each comparison preserves the corresponding method's continuation procedure, 
execution budget, and information access; Appendix~\ref{app:baseline-recipes} 
provides the executed baseline configurations. \textbf{3. Model and training settings.} Each \method{} instance uses typed memory, action-query reading, consequence
prediction, and calibrated release, with one learned scheduling decision per task. 
The primary generative actor is Gemma 4 31B IT, and the second AppWorld comparison 
uses GPT-OSS-20B. The controller is fitted on 90 AppWorld training tasks and 6,000 
WebShop training states. WebShop additionally uses a separate 500-state calibration 
population and five fitted seeds. The $\tau^2$-bench controller is fitted on 178 training 
tasks and evaluated on the disjoint 100-task test population. Input construction, 
fitting, release semantics, and native scoring are specified in 
Appendix~\ref{app:training}; the native interfaces are described in 
Appendix~\ref{app:extensions}.


%% file: sections/06_results.tex
\subsection{Main Results}
\label{sec:results}
\begin{table*}[t]
\setlength{\tabcolsep}{7.5pt}
\centering\small
\caption{\textbf{Complete AppWorld comparisons with published agent methods.} Each actor panel covers the same 417 challenge tasks and 139 scenarios, scored by native assertions. 
$\geq$ marks returned-usage lower bounds. 
Appendices~\ref{app:baseline-recipes} and~\ref{app:gptoss} give execution budgets, full costs and paired statistics.}
\label{tab:systems}
\begin{tabular}{lrrrrrr}
\toprule
& \multicolumn{3}{c}{Gemma 4 31B IT} & \multicolumn{3}{c}{GPT-OSS-20B}\\
\cmidrule(lr){2-4}\cmidrule(lr){5-7}
Method & TGC/417 & SGC/139 & Tokens (M) & TGC/417 & SGC/139 & Tokens (M)\\
\midrule
ReAct & 191 & 48 & 64.596 & 55 & 3 & $\geq$109.503\\
Plan-and-Execute & 50 & 4 & 30.388 & 27 & 1 & $\geq$153.653\\
Full-Code-Reflexion & 97 & 20 & 13.706 & 36 & \textbf{8} & 14.035\\
IPFunCall & 1 & 0 & 38.590 & 10 & 1 & $\geq$30.458\\
CodeAct & 14 & 0 & $\geq$42.878 & 4 & 0 & $\geq$12.202\\
LLMCompiler & 0 & 0 & 2.869 & 0 & 0 & $\geq$1.889\\
\textbf{\method{}} & \textbf{276} & \textbf{64} & $\geq$68.335 & \textbf{67} & 4 & $\geq$149.434\\
\bottomrule
\end{tabular}
\end{table*}

\begin{table}[t]
\centering\small
\setlength{\tabcolsep}{4pt}
\caption{\textbf{WebShop performance on all 500 test tasks.} \method{} uses typed memory, action-query reading, consequence prediction, and calibrated release (Eq.~\ref{eq:selection}); its training-selected controller is reported. Tokens and calls describe selected deployment. BC/PG tokens count policy inputs and calls count forward passes; LATS and \method{} count language-model input/output tokens and requests. LATS uses native search reward. Appendices~\ref{app:webshop} and~\ref{app:resources} give execution and acquisition costs.}
\label{tab:shopping}
\begin{tabular*}{\columnwidth}{@{\extracolsep{\fill}}lrrr@{}}
\toprule
Method & Score & Tokens (M) & Calls (k)\\
\midrule
Rule  & 45.29 & 0 & 0\\ 
BC & 47.22 & 9.259 & 16.879\\
PG-Transformer & 50.72 & 1.632 & 2.380\\
PG-RNN & 51.26 & 0.681 & 1.500\\
BC+PG & 49.65 & 3.238 & 5.035\\
LATS  & \textbf{77.98} & 1,076.394 & 222.822\\ 
\textbf{\method{}} & 74.69 & 0.222 & 0.377\\
\bottomrule
\end{tabular*}
\end{table}



\begin{table}[t]
\centering\small
\setlength{\tabcolsep}{4pt}
\caption{\textbf{$\tau^2$-bench on the same 100 held-out tasks.} The population contains 20 airline, 40 retail and 40 telecom tasks. Native success counts retain failed trajectories. Each row gives its executed step allowance.}
\label{tab:tau2-main}
\begin{tabular*}{\columnwidth}{@{\extracolsep{\fill}}lrrr@{}}
\toprule
Method & Successes & Rate (\%) & Max. steps\\
\midrule
LLMAgent  & 41/100 & 41.0 & 30\\ 
ReAct  & 36/100 & 36.0 & 30\\ 
\textbf{\method{}} & \textbf{59/100} & \textbf{59.0} & 200\\
\bottomrule
\end{tabular*}
\end{table}

\textbf{1. AppWorld.} On AppWorld, \method{} substantially improves both task-level and scenario-level
completion under the native execution interface (Table~\ref{tab:systems}). With 
Gemma 4 31B IT, \method{} completes 276 of 417 tasks and 64 of 139 scenarios, 
compared with 191 tasks and 48 scenarios for ReAct. The improvement remains under 
the independently fitted GPT-OSS-20B controller, where \method{} achieves the 
highest task completion among the compared methods. These results show that 
scheduling when to use source evidence can improve end-to-end application outcomes, 
rather than only changing the internal execution trace. The consistent gains across 
two generative actors suggest that the benefit comes from the compiler's release 
decision and its interaction with native execution. \textbf{2. WebShop.} On WebShop, \method{} reaches a Task Score of 74.69 on all 500 test tasks
(Table~\ref{tab:shopping}). This score is close to the stronger search-based LATS 
baseline, while \method{} uses substantially fewer selected-deployment tokens. The 
result shows that \method{} can preserve high task quality without repeatedly 
expanding the search process at deployment time. Its consequence predictions allow 
the controller to choose when additional information or preparation is worth the 
associated execution work. Thus, \method{} provides a more compact 
quality--resource operating point for native shopping tasks. \textbf{3. $\tau^2$-bench.} On $\tau^2$-bench, \method{} achieves the strongest native dialogue success among the
compared systems on the same held-out test population (Table~\ref{tab:tau2-main}). 
The evaluation covers airline, retail, and telecom tasks, while the controller is 
fitted on a separate training population. This result indicates that the scheduling 
formulation extends beyond application-state updates and shopping actions to 
multi-step conversations with different native environments. More broadly, it 
suggests that the compiler can apply the same release principle across heterogeneous 
task domains when the executor provides a suitable scheduling boundary. The additional investment required to fit the consequence model and collect candidate
continuations is reported separately from selected-deployment work 
(Appendix~\ref{app:resources}). Once the model is fitted, the deployed scheduler 
incurs only the work associated with the continuation it selects.

%% file: sections/07_analysis.tex

\subsection{Ablation Study}
\label{sec:analysis}

At each scheduling boundary, \method{} selects among ACT, WAIT, and SPEC; we compare this adaptive release with fixed-mode controls and a direct-utility controller without full consequence-coordinate supervision. In TravelPlanner, adaptive release retains the native completion rate while using less time than fixed speculation (Appendix~\ref{app:release-controls}). In WebShop, consequence-coordinate supervision improves utility over fixed SPEC and reduces selected native work (Appendix~\ref{app:ws-joint-fixed}; Tables~\ref{tab:ws-time-support} and~\ref{tab:ws-work-support}). These controls show that the learned policy can choose a useful release time and that explicit consequence prediction supports efficient continuations.

\subsection{Efficiency Study}

This study measures native work after scheduling, the value of consequence-coordinate supervision, and response to resource prices. In WebShop, adaptive release reduces selected operations and continuation tokens relative to fixed SPEC, while consequence-coordinate fitting improves value prediction and selected work under the same fitted controllers (Appendix~\ref{app:ws-joint-fixed}; Tables~\ref{tab:ws-time-support} and~\ref{tab:ws-work-support}). Acquisition and fitting costs are separated from recurring selected-deployment work in Appendix~\ref{app:resources}; the full price-response uncertainty family is retained in Appendix~\ref{app:sensitivity}.

\subsection{Generalization Across Execution Interfaces}

We apply the same release decision to ScienceAgentBench, where the system can generate immediately, wait for a data profile, or prepare a reversible function skeleton. Adaptive release reduces generation and replaced-completion work relative to fixed SPEC (Appendix~\ref{app:science}), showing that consequence-based scheduling transfers beyond WebShop.

\subsection{Consequence-Coordinate Supervision Reduces Native Work}
\label{sec:consequence-work}

Using the same actor, task population, fitting seeds, and calibration procedure, consequence-coordinate supervision reduces selected time and native work relative to direct-utility supervision, while the same adaptive controller reduces selected work relative to fixed SPEC. Explicit quality and resource coordinates improve selected deployment even when the actor and calibration protocol are held fixed, because the adaptive rule prepares a prefix only when its predicted continuation benefit justifies that cost; acquisition and all-attempt costs remain in Appendix~\ref{app:resources}. Figure~\ref{fig:consequence-work} summarizes these matched comparisons.

\begin{figure}[!t]
\centering
\includegraphics[width=\columnwidth]{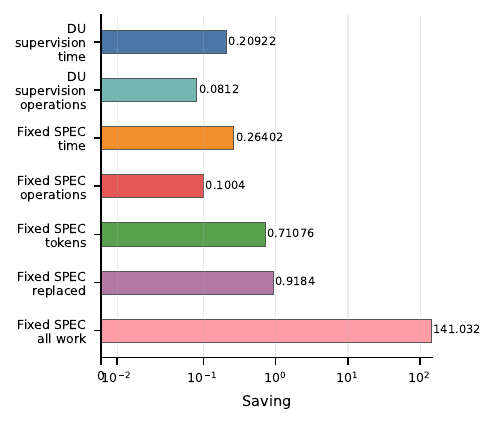}
\caption{\textbf{Matched WebShop work savings.} Bars show control minus the full controller for the 500-task population and five retained fits. Labels retain the native endpoint units; the logarithmic axis is descriptive because time, native operations and continuation tokens have different scales.}
\label{fig:consequence-work}
\label{tab:ws-time-support}\label{tab:ws-work-support}
\end{figure}

\subsection{Resource-Price Adaptation Changes Release Behavior}
\label{sec:resource-price}

\begin{figure}[!t]
\centering
\includegraphics[width=.9\columnwidth]{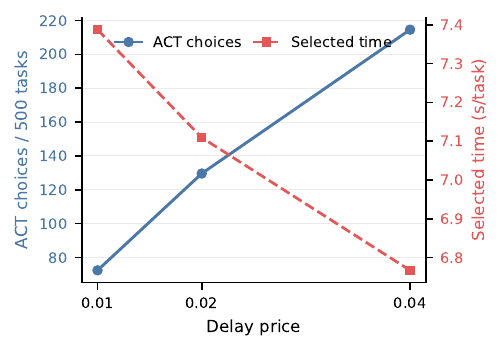}
\vspace{-1em}
\caption{\textbf{WebShop response to delay price.} The two lines share the same delay-price axis; ACT choices use the left scale and selected deployment time uses the right scale. Points average the five retained fits over all 500 tasks.}
\label{fig:ws-delay-price}
\label{tab:ws-within-price}
\vspace{-1em}
\end{figure}

\begin{figure}[!t]
\centering
\includegraphics[width=\columnwidth]{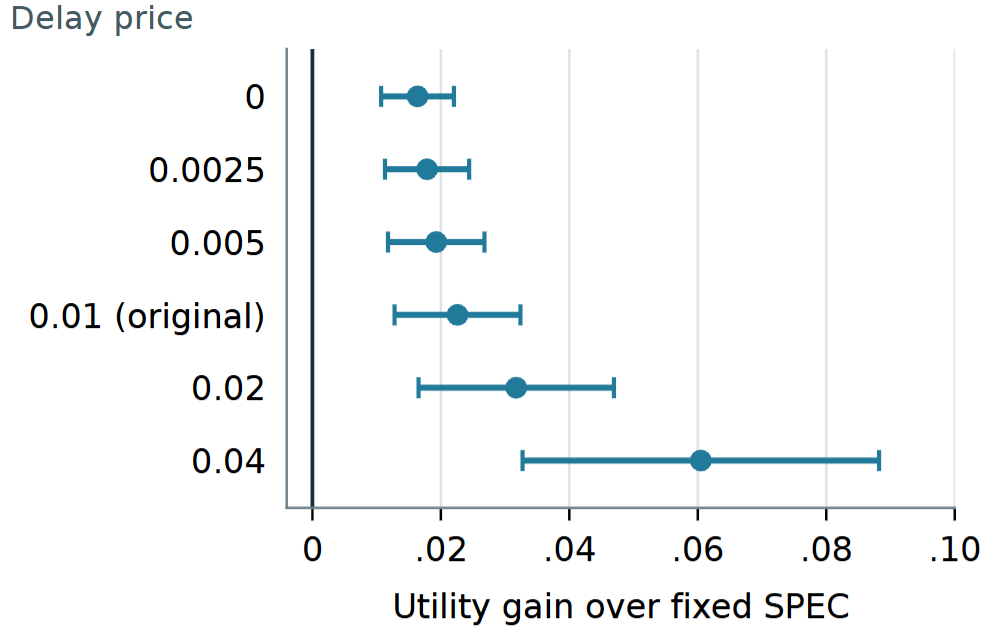}
\caption{\textbf{One consequence readout, six delay prices.} Utility gains over fixed SPEC use all 500 native WebShop tasks and 20 initializations. Each fit reselects actions from fixed predictions with its original calibration margin. Bars are joint empirical fit--task simultaneous 95\% intervals over all 18 fixed-action--price contrasts. Both policies are valued at the displayed price.}
\label{fig:ws-initialization-price}
\end{figure}

Holding consequence weights and calibration margins fixed, repricing the delay term shifts the controller toward ACT and lowers selected deployment time and native work. The action-share shift connects a resource preference to scheduling behavior and execution efficiency; changing the price vector changes the utility comparison without refitting. Figure~\ref{fig:ws-delay-price} gives the operating points, while Figure~\ref{fig:ws-initialization-price} shows that the same consequence readout retains positive utility gains across the recorded price grid; complete uncertainty families remain in Appendix~\ref{app:sensitivity}.

\begin{figure}[!t]
\centering
\includegraphics[width=.9\columnwidth]{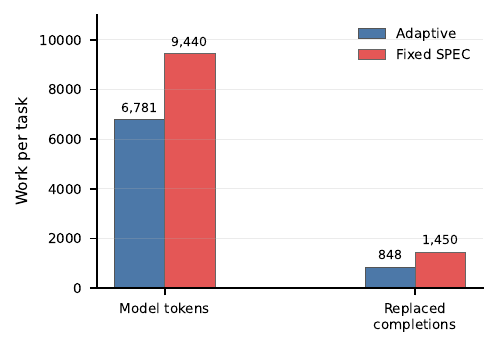}
\vspace{-1em}
\caption{\textbf{ScienceAgentBench program-generation work.} Narrow bars compare adaptive and fixed SPEC generation on 100 tasks in 69 data-directory groups. Values average five fitting initializations; savings are fixed SPEC minus adaptive.}
\label{fig:science-program-work}
\label{tab:science-resource-support}
\vspace{-1em}
\end{figure}

\subsection{Adaptive Release Reduces Scientific Program Preparation}
\label{sec:science-preparation}

At the program-generation boundary, adaptive release reduces model-generation and replaced-completion work relative to fixed SPEC across the ScienceAgentBench tasks and data-directory groups. ACT generates with current evidence, WAIT uses the delivered profile, and SPEC prepares a reversible function skeleton before profile-conditioned completion; predicting these consequences lets one release rule choose preparation for the current task and transfers the mechanism beyond WebShop. Figure~\ref{fig:science-program-work} reports the paired resource endpoints.

%% file: sections/08_conclusion.tex
\section{Conclusion}
In this work, we bring soft dependencies into the scheduling of tool-using agents. Hard dependencies establish execution readiness; soft dependencies make the value of starting depend on evidence that can improve the continuation. \method{} learns the quality and resource consequences of immediate execution, waiting, and reversible preparation, then compiles calibrated legal utility into a native execution directive. Complete application, shopping and held-out conversational populations establish the resulting task outcomes through external method comparisons; a second application actor adds an independently fitted comparison. Shared-state schedules, controlled memory reading, and consequence-supervision studies connect the neural core to prepared work, terminal completion, and resource-price response. Repeated fitting, distinct-input sensitivity, and grouped program-generation outcomes extend these observations across initializations, public states, and native execution interfaces. This formulation gives agent compilation a learned way to coordinate executable work with the information that makes it useful.

%% file: sections/09_limitations.tex
\section*{Limitations}
The measured \method{} instances make one learned scheduling decision per task at a designated soft dependency. The controller does not perform global dependency discovery or jointly schedule multiple pending sources and executable operations; those choices remain with the native interface or a higher-level planner. Repeated release boundaries require training trajectories that cover successive observations and commitments. Actor-specific fitting follows the native consequence distribution of the model executing those decisions; Appendix~\ref{app:training} specifies each measured training population and recipe.

The native interfaces provide an executable prefix whose externally visible effects can be held until observation. Deployment to external services requires authorized state changes, a reliable commit boundary and the corresponding outcome feedback. Recorded client time follows the stated execution budgets and serving conditions; the resource tables retain selected-deployment and acquisition scopes.

%% file: appendix/teaser_candidates_20261008.tex

%% file: appendix/a_training.tex
\section{Neural Compiler and Native Execution Recipes}
\label{app:training}
Each \method{} recipe starts from the native ready-state interface, builds typed memory, reads it with action queries, and maps predicted consequences to a calibrated legal execution directive. Input dimensions, normalization, targets and weights adapt those stages to the native environment. Fixed actor weights execute the matched branches used for controller fitting and the selected continuation at deployment.

\subsection{Common notation}
\label{app:symbols}
Table~\ref{tab:symbols} links the compiler's state, neural computation, and execution directive. Numeric dimensions and optimization settings are instantiated by the measured recipes below.
\begin{table*}[!t]
\centering\small
\caption{\textbf{Shared compiler notation.} The same symbols describe every native interface and fitted instance.}
\label{tab:symbols}
\begin{tabularx}{\textwidth}{@{}lX@{}}
\toprule
Symbol & Meaning\\
\midrule
$x,P,L,b$ & User request, remaining agent program, observed execution ledger, and resource budget\\
$\omega,\phi(\omega),o$ & Information event, its public description, and its returned observation\\
$s^-,s^+$ & Visible state before and after the designated observation\\
$\mathcal A,a,m(a),a^*$ & Execution modes, one mode, native legality mask, and selected directive\\
$X_N,X_E,x_R$ & Operation rows, information-event rows, and resource row\\
$N,E,d_N,d_E,d_R$ & Row counts and native input widths\\
$d,n_h,d_h,d_{\rm ff}$ & Memory width, attention-head count, head width, and feed-forward hidden width\\
$E_k,e_k,Z,A$ & Type encoder, type vector, combined memory, and three learned action queries\\
$C,H_1,H_2$ & Attention output and the two residual action representations\\
$B,\nu_Q,\nu_K,\gamma_h$ & Memory validity mask, query/key normalization, and head scale\\
$y_a,\widehat Y_a,q_a,c_{aj}$ & Observed/predicted consequences, quality, and resource coordinate $j$\\
$U_a,\alpha,\lambda_j,\kappa_a$ & Mode utility, quality weight, resource price, and calibration offset\\
$\Delta_{\wait}(s^-)$ & Expected WAIT-minus-ACT utility at the scheduling boundary\\
$M_{iaj},\delta,\mathcal L_{\rm con}$ & Observed-target mask, Huber transition, and consequence loss\\
\bottomrule
\end{tabularx}
\end{table*}

\subsection{Measured layers and parameter accounting}
\label{app:layers}
All three primary instances use $d=32$, three learned queries, $n_h=4$ heads with $d_h=8$, two residual updates, $d_{\rm ff}=64$, and five output coordinates per action. Table~\ref{tab:parameters} accounts for their stored parameters.

TravelPlanner and AppWorld type encoders apply input LayerNorm, an affine map to width 32, SiLU, and a second affine map. TravelPlanner uses scaled dot-product attention, with $\nu_Q=\nu_K$ equal to the identity and $\gamma=1/\sqrt{8}$ in Eq.~\ref{eq:attention}. AppWorld normalizes each query and key head to unit $\ell_2$ norm. Its four positive scales start at $\sqrt{8}$ and use learned log-scales capped at 100. Both use separate residual LayerNorms, SiLU in the feed-forward map, and a linear quality readout with softplus cost readouts.

WebShop uses affine--GELU--LayerNorm type encoders and normalized queries and keys. Its Q/K/V projections are bias-free; the output projection is affine, and four learned positive scales start at one. The two residual updates share LayerNorm parameters. A further LayerNorm precedes the final linear five-coordinate map, so the two stored normalization modules contain 128 parameters. The output represents standardized consequences. This parameterization preserves the stage order in Eqs.~\ref{eq:memory}--\ref{eq:prediction}. Controller dropout is zero in all three instances.

\begin{table*}[!t]
\centering\small
\caption{\textbf{Measured controller parameters.} Input widths are operation/event/resource widths; WebShop uses the corresponding product/goal/statistic types. The fixed MPNet encoder is accounted for separately. The stored auxiliary branch in TravelPlanner and AppWorld has zero coefficient in both fitting and selection.}
\label{tab:parameters}
\begin{tabular}{@{}lrrr@{}}
\toprule
Component & TravelPlanner & AppWorld & WebShop\\
\midrule
Input widths & 57/24/51 & 35/22/6 & 768/768/9\\
Three type encoders & 7,752 & 5,406 & 49,728\\
Type vectors and action queries & 192 & 192 & 192\\
Attention projections and scales & 4,224 & 4,228 & 4,132\\
Residual/readout normalization & 128 & 128 & 128\\
Feed-forward map & 4,192 & 4,192 & 4,192\\
Linear consequence map & 165 & 165 & 165\\
Stored auxiliary branch & 1,089 & 1,089 & 0\\
\midrule
Total stored parameters & 17,742 & 15,400 & 58,537\\
Parameters in active prediction/loss path & 16,653 & 14,311 & 58,537\\
\bottomrule
\end{tabular}
\end{table*}

\begin{table*}[!t]
\centering\small
\caption{\textbf{Fitting settings of the selected compiler instances.} The paragraphs below define the target units, objectives, native actions, and calibration rules.}
\label{tab:fitting-settings}
\begin{tabular}{@{}lccc@{}}
\toprule
Setting & TravelPlanner & AppWorld & WebShop\\
\midrule
Training states & 27 & 90 & 6,000\\
Optimizer & Adam & Adam & AdamW\\
Learning rate & 0.003 & 0.003 & 0.0008\\
Weight decay & 0 & 0 & 0.01\\
Batch & Full & Full & 256\\
Training duration & 200 updates & 200 updates & 100 epochs\\
Huber transition $\delta$ & 1 & 1 & 0.25\\
Calibration & Zero offsets & Zero offsets & Separate 500 states\\
\bottomrule
\end{tabular}
\end{table*}

\subsection{TravelPlanner internal study}
\label{app:travelplanner}
TravelPlanner~\citep{xie2024travelplanner} supplies a internal study of scheduling inside a shared constraint-aware itinerary executor. All internal controls retain the same source acquisition, observed-catalog projection, construction checks, actor and native Final Pass scorer on 180 validation requests. The fixed ACT, WAIT and SPEC controls retain this execution framework with a constant release rule; the adaptive controller learns the release choice. Appendix~\ref{app:release-controls} reports these framework controls together with their complete quality and resource outcomes, and Appendix~\ref{app:components} reports matched component interventions.

\paragraph{Typed memory.}
Operation rows represent city--tool support, observed lodging, dining and attraction items, and transport legs. Event rows describe pending catalog queries through their tool identity and public arguments. Resource fields describe the party, dates, budget, requested constraints and consumed work. Request-local city slots align cities across row types. The 57 operation coordinates cover type/tool, city and endpoints, observation counts, prices, occupancy, minimum stay, compatibility, relative dates, room and cuisine constraints, and transport legality. The 24 event coordinates describe tool identity, query order, city/endpoints, public matching information, relative date and legality. The 51 resource coordinates encode the request constraints, pending queries, observed items, and source calls, tokens and seconds.

Missing numeric fields carry a known-value bit. Nonnegative values use $\log(1+v/s)$; relative dates use $\operatorname{sign}(v)\log(1+|v|)$. Monetary quantities use $s=1000$, consumed tokens use $10000$, source seconds use $60$, and absolute day-of-year uses $366$; other counts use one. Missing and observed-zero fields map to $(0,0)$ and $(0,1)$, respectively. Categorical constraints use explicit indicators.

\paragraph{Action-query reading and consequence learning.}
The layers in Appendix~\ref{app:layers} produce three five-coordinate predictions. Fitting uses 27 public training states with all 81 ACT/WAIT/SPEC outcomes. The targets are $y_1=Q$, $y_2=T/1000$, and $y_3=K/10^6$, where $Q$ is native Final Pass, $T$ is deployment seconds and $K$ is model tokens. The final two coordinates are masked. All first-three targets are observed for this training collection. The loss sums the three coordinate-wise mean Huber losses with beta one. Full-batch Adam uses 200 updates, learning rate 0.003, and zero weight decay.

\paragraph{Calibrated release.}
Utility is $U=y_1-y_2-y_3$, with zero release offsets. The selected fit assigns 126 validation requests to ACT, 14 to WAIT and 40 to SPEC. The actor is \nolinkurl{google/gemma-4-31B-it}, revision \nolinkurl{439edf5652646a0d1bd8b46bfdc1d3645761a445}. It receives public tools, observed catalog rows and construction feedback. ACT begins a draft with available facts; WAIT releases after the pending facts arrive; SPEC permits one draft while holding submission. Later requests incorporate the delivered facts. Each deployed task pays source acquisition, producer work and its selected continuation.

\paragraph{Itinerary construction and commitment.}
The executor normalizes admissible tool requests, checks the remaining budget, and lowers them to native calls. An observed-catalog projection keeps the proposed city order, transfer days, modes, lodging blocks and meal-slot cities fixed. It selects the cheapest compatible observed transfer and a room satisfying the block's city, occupancy, minimum-stay and requested constraints. Meal allocation groups variants by restaurant identity and uses dynamic programming over meal count and a cuisine bitmask. Processing an identity group extends the previous states once, ensuring distinct restaurant identities. A second dynamic program combines city allocations to cover the required cuisines at minimum observed cost.

Public checks then assess budget, continuity, lodging, cuisines and transport compatibility. The actor can revise the itinerary structure using this feedback. A speculative first response remains behind the commit boundary; subsequent responses can use the producer output. At budget exhaustion the executor exports its last complete post-prefix candidate. The native Final Pass scorer evaluates all 180 official validation requests.

\subsection{AppWorld}
\label{app:appworld}
\paragraph{Typed memory.}
Three operation rows, one event row and one resource row encode the public instruction before API execution. Each 35-coordinate operation row contains a three-way row identity, nine instruction statistics and 23 cues. The cues cover nine applications and fourteen intent patterns; training-observed cues define the fixed mask. Statistics summarize words and applications, source-note and date references, exclusions, writes, destructive operations, exports and cross-application scope. Six resource coordinates encode normalized word/application counts, writes, destructive operations, question form and exports.

The event row joins six instruction-derived coordinates to a 16-dimensional semantic projection. A fixed \nolinkurl{sentence-transformers/all-mpnet-base-v2} encoder produces 768-dimensional instruction embeddings. The training mean and first 16 training principal components define the projection. Coordinates are divided by training standard deviations floored at $10^{-3}$ and clipped to $[-5,5]$. The saved transformation applies to challenge tasks. Encoded rows have shapes $3\times35$, $1\times22$ and $6$ before the common width-32 memory.

\paragraph{Action-query reading and consequence learning.}
The AppWorld layers use normalized Q/K attention and separate residual norms as specified above. Let $S$ be task success, $P$ the fraction of native assertions passed, $T_{\rm ms}$ model milliseconds, $A_{\rm api}$ API calls and $K$ model tokens. Targets and utility are
\begin{align}
 y_1&=S+0.05P,\quad y_2=T_{\rm ms}/600000,\nonumber\\
 y_3&=A_{\rm api}/200+K/200000,\nonumber\\
 y_4&=y_5=0,\quad U=4y_1-0.1y_2-0.1y_3.
\end{align}
All 270 outcomes from 90 training tasks contribute. The loss is mean Smooth-$L_1$ over the first three coordinates, with beta one, plus twice cross-entropy over predicted utilities with the best observed action as target. Full-batch Adam uses 200 updates, learning rate 0.003, and zero weight decay. Training folds group the 30 scenarios; the selected final checkpoint fits all 90 tasks. The recorded numerical environment is Python 3.12 and PyTorch 2.11.0/CUDA 12.9.

\paragraph{Calibrated release.}
All three actions are legal at this initial decision. Inference uses zero release offsets and utility temperature one. ACT, WAIT, SPEC order resolves ties. The selected checkpoint assigns 116 tasks to ACT, 159 to WAIT and 142 to SPEC; its parameters and saved inputs reproduce all 417 choices. The fixed actor is the Gemma revision given above. It allows 100 model calls, 500 environment interactions, 1,024 output tokens per generation, and temperature zero.

\paragraph{API execution and commitment.}
ACT starts with the available evidence. WAIT incorporates a required source or business read before a dependent state change. SPEC prepares API-schema inspection and independent target reads, then incorporates the source result before committing the change. The runtime parses generated Python and observes actual API request history. A business read is an application GET outside authentication, supervisor and API-documentation routes. Instruction cues designate Simple Note content or current-date reads when required. WAIT and SPEC hold a business write until a business read and each designated source read appear in the history.

The common continuation instructions specify observation memory, completion answers, order identifiers, purchase budgets and API validation. The actor checks returned fields, record joins and output format during execution. Native assertions evaluate the final state. TGC counts successful tasks; SGC requires all three tasks in a scenario to succeed. The challenge population comprises 417 tasks and 139 scenarios.

\subsection{WebShop}
\label{app:webshop}
\paragraph{Typed memory.}
The public goal and up to 32 visible first-page product titles receive fixed 768-dimensional MPNet embeddings. Titles are case-folded; tokenization truncates at 96 tokens. Attention-mask mean pooling and $\ell_2$ normalization produce the vectors. A validity mask excludes missing products. Goal, product and scalar-statistic encoders supply the three types in the width-32 memory. The goal describes the information relevant to a prospective product lookup; the controller sees the available titles before that lookup's result is released.

For goal vector $g$, valid title vectors $p_i$ and similarities $s_i=p_i^\top g$, the nine public statistics are the two largest similarities, mean, population standard deviation, minimum, top-two gap, product count divided by 32, maximum goal--title token Jaccard overlap, and minimum displayed price divided by 1000. A single product supplies both largest values. Lexical tokens are case-folded alphanumeric spans. The first displayed decimal supplies the price, with zero for unavailable values. Scalar centering and scaling use training moments.

\paragraph{Action-query reading and consequence learning.}
The measured WebShop layers in Appendix~\ref{app:layers} predict native reward $Q$, user seconds $T$, occupied seconds, option-model tokens $K$ and native operations $A$. Training standardizes each coordinate over 6,000 states and all three actions; inference restores the units. The public-detail producer contributes $A_{\rm producer}$ operations, charged once per deployed task. Utility is
\begin{equation}
 U=Q-0.01T-10^{-4}K-10^{-3}(A+A_{\rm producer}).
\end{equation}
The loss is $0.70\mathcal L_{\rm outcome}+0.25\mathcal L_{\rm utility}+0.02\mathcal L_{\rm conservative}$. Outcome regression averages the five coordinate losses, $\mathcal L_{\rm outcome}=\mathcal L_{\rm con}/5$. Outcome and standardized-utility regression use Smooth-$L_1$ with beta 0.25. For standardized predicted utility $\widetilde U$,
\begin{equation}
\begin{split}
\mathcal L_{\rm conservative}=\frac1n\sum_i\Bigl[&
0.5\log\sum_a e^{\widetilde U_{ia}/0.5}\\
&-\frac13\sum_a\widetilde U_{ia}\Bigr].
\end{split}
\end{equation}
We use AdamW, learning rate $8\times10^{-4}$, weight decay 0.01, batch size 256, 100 epochs and gradient clipping at one. The recorded fitting environment is Python 3.13, PyTorch 2.8.0/CUDA 12.8 and NumPy 2.3.2 on a B200.

\paragraph{Calibrated release.}
Each seed selects a common ACT/SPEC offset from $\{0,0.02,0.05,0.10,0.15\}$ by mean utility on a separate 500-state calibration population; WAIT has offset zero. The selected margins are $0.05,0.02,0.15,0.02,0.02$. Strict threshold comparisons favor WAIT at equality and ACT on an ACT/SPEC tie. The 6,000 fitting states have distinct target products; calibration targets are disjoint from them. Evaluation uses official test indices 0--499. Retained parameters and normalizations reproduce all 2,500 seed--task choices.

\paragraph{Search, preparation and purchase.}
A common initial query extracts public goal keywords. The public-detail producer traverses at most five result pages, ranks visible candidates with the shared value function, and inspects detail/option pages for three distinct products. It selects the product with most goal-matched option groups, resolving ties by candidate value. All branches share these producer facts after release.

ACT ranks the first page, opens its selected product, chooses options and purchases. WAIT receives the producer result before navigating to its selected product. SPEC opens the best first-page candidate and prepares options while holding purchase. After release, it reuses a compatible option plan for the same product or returns to the prior search and selects the producer's product. The Gemma option actor receives the goal, title, displayed price and indexed option groups; it uses temperature zero, a 512-token output cap and JSON selections. A shared lexical option check merges selected values before native clicks and purchase. The native purchase reward supplies Task Score.

%% file: appendix/b_results.tex
\section{Reference Systems and Resource Accounting}
\label{app:results}
The resource records bind the selected native terminal to its actor, online information, executed budget, and complete task population. Selected-deployment work and all-attempt investment retain their recorded scopes.
\subsection{Reference execution recipes}
\label{app:baseline-recipes}
The primary reference systems use the same native populations and scorers as their corresponding \method{} instances. AppWorld generative comparisons use the Gemma revision specified in Appendix~\ref{app:travelplanner}. WebShop's released learned policies use their own policy networks; the local LATS search uses the Gemma actor.

\paragraph{AppWorld released agents.}
ReAct, Plan-and-Execute, Full-Code-Reflexion, IPFunCall and CodeAct use their released AppWorld implementations and native task assertions \citep{trivedi2024appworld}. ReAct and Plan-and-Execute permit 100 model calls and 500 environment interactions, with 400 output tokens per generation, temperature zero and seed 123. CodeAct permits 40 model calls and 40 environment interactions, with 2,048 output tokens. Full-Code-Reflexion uses 10 outer steps, five reflection retries and 4,096 output tokens. IPFunCall uses 15 steps and the same output cap. The latter two use seed 100 and three demonstrations from the 90-task training split.

LLMCompiler retains its author planner, graph parser, dependency substitution, task-fetching unit and joiner. Streaming planning permits one planning round, four concurrent generation slots, 100 model calls and 500 environment interactions. Planner, executor and joiner output caps are 8,192, 4,096 and 1,024. Each task's first valid native terminal in invocation order supplies its measured outcome. Interrupted and failed physical attempts contribute to the associated resource scope.

\paragraph{WebShop references.}
The rule, behavior-cloning (BC), transformer and recurrent policy-gradient (PG), and BC+PG policies follow the benchmark release \citep{yao2022webshop}. Their resource fields count policy input tokens and model forward passes. LATS follows its language-agent tree search with native environment feedback \citep{zhou2024lats}. Its search can observe native purchase reward. The release controller and its candidate/option procedure use public task, page and option information. The complete comparison preserves these information boundaries. Score is the mean native reward times 100 on all 500 official test indices.

\subsection{Resource definitions and complete populations}
\label{app:resources}
A model-token count sums returned prompt and completion usage. A request count records physical generation requests; for the released BC/PG policies it records neural forward calls as labeled. Native operations measure environment actions. Client request seconds include transport, queueing and response service. Task seconds sum each recorded task interval, including concurrent intervals where applicable. These elapsed quantities retain their measured scopes; GPU occupancy would require separate device measurements.

Acquisition collects the outcomes used for fitting and calibration. Selected deployment charges one chosen continuation and its shared producer per task. All-attempt accounting adds interrupted and failed physical executions within the named campaign. A lower bound sums the observed work when coverage is incomplete. Native task failures remain in the quality denominator regardless of resource coverage.

\paragraph{TravelPlanner internal execution resources.}
The selected internal workflow covers all 180 requests and counts source, producer, compiler and selected-continuation work: 9,655,472 model tokens, 1,097 requests, 11,218.94 client request seconds and 10,333.28 task seconds. Worker elapsed time is 14,091.33 seconds including setup; request and task intervals can overlap. Table~\ref{tab:fixed-tp} gives the corresponding fixed-rule framework controls.

TravelPlanner label acquisition for 27 states and 81 continuations uses 4,356,380 model tokens, 360 requests, 624 physical tool queries and 15,465.32 client request seconds. Source and producer acquisition are charged once across the three branches. Source-grouped selection and final fitting take 38.33 seconds with 23.39 MiB peak allocated GPU memory; training-outcome scoring takes 13.73 seconds. Native scoring for the selected validation deployment takes 31.73 seconds separately from execution.

\begin{table*}[!t]
\centering\small
\caption{\textbf{AppWorld resources associated with the complete Gemma comparisons.} Each row binds native task completions and resources to the same campaign. Time is task-summed kiloseconds; CodeAct uses a complete timed replication. Counts include all physical attempts. Token lower bounds retain missing-response coverage.}
\label{tab:aw-cost}
\begin{tabular}{@{}lrrrr@{}}
\toprule
System & TGC/417 & Tokens (M) & Requests & Task time (ks)\\
\midrule
ReAct & 191 & 64.596301 & 8,668 & 127.164\\
Plan-and-Execute & 50 & 30.387994 & 4,362 & 112.646\\
Full-Code-Reflexion & 97 & 13.706327 & 1,548 & 78.542\\
IPFunCall & 1 & 38.590123 & 5,959 & 93.066\\
CodeAct (timed) & 13 & $\geq$45.341099 & 7,108 & 136.174\\
LLMCompiler & 0 & 2.868727 & 2,252 & 33.843\\
\method{} & 276 & $\geq$68.334974 & 8,177 & 110.651\\
\bottomrule
\end{tabular}
\end{table*}

The AppWorld controller's total covers 456 complete and interrupted executions, with 8,169 response events, six exceptions and two cancellations. Measured response elapsed time sums to 94,338.10 seconds. The complete LLMCompiler campaign additionally includes prerequisite and canary attempts; its client request time is 35,799.19 seconds. The quality panel reports CodeAct's original 14 completions, with 42,877,586 selected-terminal tokens and 6,944 known-attempt calls. Its complete timed replication uses the same actor revision, generation settings, native scorer and 40-call execution budgets. Across all 417 tasks it records 13 completions, 420 physical attempts, 7,108 requests and 136,173.57 task seconds; client requests sum to 125,389.90 seconds. One timed-out request has unknown token usage, giving the reported token lower bound. Pairing every task ID in the complete timing cohort with the selected \method{} deployment, \method{} completes 276/417 tasks and all three tasks in 64/139 scenarios, giving gains of 63.07 TGC points and 46.04 SGC points over timed CodeAct. The joint simultaneous 95\% intervals are $[55.16,70.98]$ and $[35.86,56.22]$, with Holm-adjusted $p=1.12\times10^{-32}$ and $3.25\times10^{-19}$, respectively. These intervals retain one four-endpoint family containing TGC, SGC, all-attempt task time and physical request counts. Paired inference resamples whole scenarios for 50,000 draws (seed 20261008); quality tests use exact scenario sign flips, resource tests use 50,000 scenario sign flips (seed 20261009), and Holm spans all four tests. The contrasts describe the executed workflows under their reported budgets and serving conditions.

\begin{table*}[!t]
\centering\small
\caption{\textbf{WebShop learning and recurring deployment resources.} Collection covers all three branches. The five-fit mean and single selected controller each charge the shared producer once. Client seconds sum model requests; encoding seconds record the associated text encoding. Dashes mark timing fields outside the single-controller receipt.}
\label{tab:ws-learning}
\begin{tabular}{@{}lrrrrrr@{}}
\toprule
Scope & States & Tokens & Requests & Native steps & Client s & Encoding s\\
\midrule
Training and calibration & 6,500 & 8,146,299 & 15,981 & 178,179 & 74,787.04 & 15,415.14\\
Three-action test collection & 500 & 700,403 & 1,241 & 13,724 & 8,112.21 & 941.60\\
Selected deployment (five-fit mean) & 500 & 216,928.2 & 373.8 & 9,096.8 & 2,539.87 & 725.22\\
Selected deployment (single controller) & 500 & 221,952 & 377 & 9,045 & -- & --\\
\bottomrule
\end{tabular}
\end{table*}

The single-controller row corresponds to the selected WebShop controller evaluated on all 500 official test tasks in the main results. Its selected deployment records 218,579 prompt tokens, 3,373 completion tokens, 377 physical requests, 9,045 native operations including the shared producer, and 4,990.64 user seconds. The five-fit mean row summarizes the earlier five fitted controllers and retains its original timing scope.

Initial public-vector construction takes 42.74 seconds. Fitting and calibration of the five selected WebShop controllers take 102.07 seconds. The 20-initialization study in Appendix~\ref{app:initialization} comprises 80 fits across four methods; the 60 additional fits use the retained acquisition data and require 3,565.82 aggregate preparation-and-fitting seconds. The combined collection uses 8,846,702 model tokens and 17,222 physical requests, with complete returned usage. Mean selected deployment takes 3,693.67 user seconds across the 500 tasks, including the native operations and shared producer.

Per-fit preparation times, GPU memory peaks and selected calibration margins are retained in the recorded fitting accounting. The four original workflows, each covering five fits, retain their full reconstruction or control workflow times in the same accounting.

\paragraph{Per-fit preparation and memory.}
The additional initialization runs give a concrete resource distribution for fitting the shared controller on cached inputs (Table~\ref{tab:ws-fit-resources}). Each of four methods has 15 complete fits, using the retained 6,000 fitting, 500 calibration, and 500 test inputs. The full controller has 58,537 trainable parameters. Wall time covers preparation and fitting for each recorded workflow; CUDA memory is its peak allocated memory. Acquisition and selected-deployment resources are reported above. The four original five-fit workflows retain their separately recorded reconstruction scope.

\begin{table}[tbp]
\centering\footnotesize
\setlength{\tabcolsep}{3pt}
\caption{Additional WebShop initialization-fit resources: 15 fits per row, 60 in total. Wall times use cached inputs and are in seconds; peak allocated CUDA memory is in MiB. Each row reports the median, observed range, and largest memory peak.}
\label{tab:ws-fit-resources}
\begin{tabularx}{\columnwidth}{@{}Xrrr@{}}\toprule
Controller & Median & Range & Peak\\\midrule
\method{} & 80.33 & 38.62--92.26 & 636.89\\
Uniform attention & 87.06 & 45.20--99.00 & 636.89\\
Utility-only target & 44.03 & 25.61--60.24 & 636.89\\
Pooled MLP & 40.53 & 23.25--47.15 & 628.25\\
\bottomrule
\end{tabularx}
\end{table}

For the released rule, BC, PG-Transformer, PG-RNN and BC+PG comparisons, task-summed inference times are 0.18, 3.26, 0.28, 0.16 and 0.51 kiloseconds, respectively. Their policy-input and forward-call counts are in Table~\ref{tab:shopping}. LATS selected deployment uses 1,076,393,717 tokens, 222,822 requests, 303,215 environment attempts and 2,688,689.25 task seconds. Across its wider 773-attempt record, the measured lower bounds are 1,916,871,437 tokens, 394,411 request starts and 3,807,260.76 task seconds; 25 compressed request streams have incomplete coverage. The selected-deployment comparison and the all-attempt investment therefore have explicitly different resource scopes.

\begin{table}[!t]
\centering\small
\setlength{\tabcolsep}{4pt}
\caption{\textbf{Completed CodeAct retest on AppWorld.} The GPT-OSS-20B retest of CodeAct~\citep{wang2024codeact} covers all 417 challenge tasks and 139 scenarios, with same-response action decoding and native execution feedback. All 417 terminal database states were independently re-evaluated with the native AppWorld assertion scorer. Task failures remain in the denominator. Retained attempts record at least 12.202 million model tokens and 4,213 generation requests.}
\label{tab:appworld-codeact-retest}
\begin{tabular*}{\columnwidth}{@{\extracolsep{\fill}}lrrr@{}}
\toprule
Method & TGC/417 & SGC/139 & Rechecked\\
\midrule
CodeAct & 4 & 0 & 417/417\\
\bottomrule
\end{tabular*}
\end{table}

\begin{table}[!t]
\centering\small
\setlength{\tabcolsep}{4pt}
\caption{\textbf{Completed IPFunCall retest on AppWorld.} The GPT-OSS-20B retest of the released IPFunCall agent~\citep{trivedi2024appworld} covers all 417 challenge tasks and 139 scenarios. Public API selection and function calling retain the registered native execution limits. All 417 terminal database states were independently re-evaluated with the native AppWorld assertion scorer. Retained attempts record at least 30.458 million model tokens and 6,141 generation requests.}
\label{tab:appworld-ipfuncall-retest}
\begin{tabular*}{\columnwidth}{@{\extracolsep{\fill}}lrrr@{}}
\toprule
Method & TGC/417 & SGC/139 & Rechecked\\
\midrule
IPFunCall & 10 & 1 & 417/417\\
\bottomrule
\end{tabular*}
\end{table}

\begin{table}[!t]
\centering\small
\setlength{\tabcolsep}{4pt}
\caption{\textbf{Completed LLMCompiler retest on AppWorld.} The GPT-OSS-20B retest instantiates the released planner, dependency execution and joiner stages~\citep{kim2024llmcompiler} on all 417 challenge tasks and 139 scenarios. All 417 terminal database states were independently re-evaluated with the native AppWorld assertion scorer. Retained attempts record at least 65.414 million model tokens and 15,439 generation requests. Compatible retained histories preserve effective planner/executor text and author-parsed joiner decisions.}
\label{tab:appworld-llmcompiler-retest}
\begin{tabular*}{\columnwidth}{@{\extracolsep{\fill}}lrrr@{}}
\toprule
Method & TGC/417 & SGC/139 & Rechecked\\
\midrule
LLMCompiler & 0 & 0 & 417/417\\
\bottomrule
\end{tabular*}
\end{table}

\begin{table}[!t]
\centering\small
\setlength{\tabcolsep}{4pt}
\caption{\textbf{Completed Gemma IPFunCall retest on AppWorld.} The Gemma-4-31B-IT retest of the released IPFunCall agent~\citep{trivedi2024appworld} covers all 417 challenge tasks and 139 scenarios. Public API selection and function calling retain the registered execution limits. All 417 terminal database states were independently re-evaluated with the native AppWorld assertion scorer. Retained attempts record at least 51.258 million model tokens.}
\label{tab:appworld-ipfuncall-gemma-retest}
\begin{tabular*}{\columnwidth}{@{\extracolsep{\fill}}lrrr@{}}
\toprule
Method & TGC/417 & SGC/139 & Rechecked\\
\midrule
IPFunCall & 7 & 0 & 417/417\\
\bottomrule
\end{tabular*}
\end{table}

%% file: appendix/c_statistics.tex
\section{Paired Inference and Controlled Comparisons}
\label{app:statistics}
The controlled studies follow \method{} from soft-dependency scheduling to memory reading, consequence supervision, and resource-price response. Every comparison retains its complete paired population, statistical family, native endpoint, and operating conditions.
\subsection{Complete native quality comparisons}
All main task vectors are joined by exact native task ID. AppWorld scenario IDs are the task-ID prefix before its final underscore; each scenario contributes exactly three tasks. WebShop uses official indices 0--499.

The AppWorld analyses retain six reference comparisons per endpoint. For a contrast family with paired differences $d_{ij}$, we resample the same units jointly across all $j$ for 50,000 bootstrap draws, with seed 20261008. AppWorld TGC uses scenario sums divided by three; SGC uses each scenario's all-three-complete indicator. Let $\bar d_j$ be the observed mean and $\widehat\sigma_j$ the bootstrap standard deviation. The 95th percentile of
\begin{equation}
 \max_j\left|\frac{\bar d_j^*-\bar d_j}{\widehat\sigma_j}\right|
\end{equation}
supplies the critical value for simultaneous intervals. The simultaneous intervals preserve the complete original comparison family. Binary task contrasts use exact McNemar tests, equivalent to exact sign flips on the discordant tasks. AppWorld tests enumerate the conditional distribution of signed integer scenario differences by dynamic programming. Holm correction spans the six tests per endpoint. These retrospective intervals describe population-unit variation conditional on the measured systems and their selection history.

\begin{table*}[!t]
\centering\small
\caption{\textbf{AppWorld TGC differences on the complete population.} Effects are \method{} minus each reference in percentage points. Intervals cover all six contrasts; tests receive Holm correction.}
\begin{tabular}{@{}lrrcr@{}}
\toprule Reference & Successes & Gain (pp) & Simultaneous 95\% CI & Holm $p$\\
\midrule
ReAct & 191/417 & 20.38 & $[12.47,28.30]$ & $2.39\times10^{-9}$\\
Plan-and-Execute & 50/417 & 54.20 & $[46.40,61.99]$ & $4.62\times10^{-33}$\\
Full-Code-Reflexion & 97/417 & 42.93 & $[33.73,52.12]$ & $3.46\times10^{-20}$\\
IPFunCall & 1/417 & 65.95 & $[58.28,73.61]$ & $3.61\times10^{-35}$\\
CodeAct & 14/417 & 62.83 & $[55.09,70.56]$ & $9.63\times10^{-35}$\\
LLMCompiler & 0/417 & 66.19 & $[58.50,73.88]$ & $3.61\times10^{-35}$\\
\bottomrule\end{tabular}\end{table*}

\begin{table*}[!t]
\centering\small
\caption{\textbf{AppWorld SGC differences on the complete population.} Effects are \method{} minus each reference in percentage points. Intervals cover all six contrasts; tests receive Holm correction.}
\begin{tabular}{@{}lrrcr@{}}
\toprule Reference & Successes & Gain (pp) & Simultaneous 95\% CI & Holm $p$\\
\midrule
ReAct & 48/139 & 11.51 & $[2.38,20.64]$ & 0.0052\\
Plan-and-Execute & 4/139 & 43.17 & $[33.15,53.18]$ & $5.20\times10^{-18}$\\
Full-Code-Reflexion & 20/139 & 31.65 & $[20.75,42.56]$ & $7.78\times10^{-10}$\\
IPFunCall & 0/139 & 46.04 & $[35.97,56.12]$ & $6.51\times10^{-19}$\\
CodeAct & 0/139 & 46.04 & $[35.97,56.12]$ & $6.51\times10^{-19}$\\
LLMCompiler & 0/139 & 46.04 & $[35.97,56.12]$ & $6.51\times10^{-19}$\\
\bottomrule\end{tabular}\end{table*}

The five selected WebShop controllers have native Task Scores 73.8310, 74.0817, 73.6393, 73.6850 and 73.7467 on the same 500 tasks. Their mean is 73.7967. Task-paired inference conditions on these fitted policies, the native purchase scorer and the recorded execution recipes.

\begin{table*}[!t]
\centering\small
\caption{\textbf{Internal TravelPlanner framework and scheduling ablation on 180 requests.} Each row shares the constraint-aware NAC itinerary executor, actor, upstream observations, construction procedure and native scorer. Fixed-rule controls retain the framework and prescribe one release mode; adaptive scheduling uses the learned controller.}
\label{tab:fixed-tp}
\begin{tabular*}{\textwidth}{@{\extracolsep{\fill}}lrrr@{}}
\toprule Framework / release rule & Pass/180 & Time (ks) & Tokens (M)\\
\midrule
NAC executor + fixed ACT & 129 & 17.395 & 9.526\\
NAC executor + fixed WAIT & 124 & 14.288 & 9.333\\
NAC executor + fixed SPEC & 128 & 64.204 & $\geq$9.724\\
NAC executor + learned scheduling & 125 & 10.333 & 9.655\\
\bottomrule
\end{tabular*}
\end{table*}

\subsection{Fixed release with a shared continuation}
\label{app:release-controls}
\paragraph{TravelPlanner framework controls.}
\label{app:tp-internal}
TravelPlanner fixed controls reuse the selected policy's pre-result state, upstream observations and downstream construction procedure. Each policy is scored on all 180 request IDs under the same constraint-aware itinerary executor. Fixed ACT is the immediate-release framework control: it retains catalog projection, budget and compatibility checks, and itinerary construction while applying ACT to every task. Fixed WAIT and SPEC retain the same framework with their corresponding constant release rule. Table~\ref{tab:fixed-tp} compares these controls with adaptive scheduling and reports complete quality and recorded deployment time. Adaptive release completes 125/180 requests in 10,333.28 task-summed seconds. It saves 299.28 seconds per task relative to fixed SPEC, with paired 95\% interval $[261.46,339.44]$. One SPEC request lacks returned usage, so its token total is the known lower bound $\geq$9.724 million.

Time savings over ACT, WAIT and SPEC are 39.23, 21.97 and 299.28 seconds per task, with pointwise 95\% intervals $[35.46,43.11]$, $[18.95,25.10]$ and $[261.46,339.44]$. The percentile bootstrap uses 20,000 task-paired draws and seed 20260927. The evaluations record different client service distributions: selected downstream requests have median 15.07 seconds, fixed ACT 21.00 seconds and fixed WAIT 17.43 seconds. The intervals characterize these elapsed-time measurements under the recorded serving conditions. The complete native pass counts and elapsed times in Table~\ref{tab:fixed-tp} characterize the shared execution framework at each release rule.

\paragraph{WebShop fixed-release controls.}
WebShop's fixed controls share the public-detail producer, option actor, native purchase scorer and all 500 task IDs. The adaptive five-fit mean Task Score is 73.7967. The paired studies below quantify how state-conditioned selection changes utility, native operations, continuation tokens and recorded duration under that common execution interface.

\paragraph{Joint fitting and task variation for fixed speculative release.}
\label{app:ws-joint-fixed}
At the same selected WebShop operating point, the adaptive-minus-SPEC utility effect is $+0.023703$, with joint simultaneous 95\% interval $[0.013514,0.033892]$; recorded time saving is 0.710762 seconds per task, with interval $[0.220673,1.200851]$. All 500 native task IDs and five fits contribute. The fits share 6,000 training states, a separate 500-state calibration population, the original 100-epoch schedule and the calibration-selected release rule. Recorded task work includes the selected native continuation and one shared producer.

For these intervals, we independently resample the five fitting seeds and 500 task IDs, applying each draw's indices jointly to every policy. We use 20,000 draws with generator seed 20261008. For each endpoint, the simultaneous critical value is the 95th percentile of the maximum absolute standardized centered bootstrap deviation across the complete three-contrast family against fixed ACT, WAIT and SPEC. The resulting intervals describe empirical fitting and task variation conditional on the retained native branches and fitting recipes. Two-sided whole-task sign flips use 50,000 draws, seed 20261009, a plus-one correction and Holm adjustment over the same three controls. Conditional on the retained five fits, the adjusted $p$ values for the utility and time contrasts against SPEC are $6.00\times10^{-5}$ and $2.00\times10^{-4}$. This controlled comparison supports adaptive selection's utility and preparation-cost advantage over always speculating.

\begin{table}[!t]
\centering\small
\setlength{\tabcolsep}{3pt}
\caption{\textbf{Adaptive release improves utility and reduces selected work relative to fixed SPEC.} All 500 WebShop tasks and five fitted controllers contribute at the original calibrated operating point. Each endpoint retains the complete three-fixed-release family, with joint seed--task simultaneous intervals and conditional Holm-adjusted tests. Utility gain is adaptive minus SPEC; time saving is SPEC minus adaptive in seconds per task.}
\label{tab:ws-joint-fixed}
\begin{tabular*}{\columnwidth}{@{\extracolsep{\fill}}lrc@{}}
\toprule
Endpoint & Effect & Simultaneous 95\% CI\\
\midrule
Utility gain & 0.023703 & $[0.013514,0.033892]$\\
Time saving & 0.710762 & $[0.220673,1.200851]$\\
\bottomrule
\end{tabular*}
\end{table}

\subsection{Memory and supervision interventions}
\label{app:components}
The claim-facing component evidence uses matched native-work endpoints with the shared actor, scorer, task population, fitting seeds and calibration procedure. Figure~\ref{fig:consequence-work} summarizes the complete positive deployment-time and native-work effects. Consequence-coordinate fitting saves 0.20922 seconds and 0.0812 native operations per task relative to direct-utility supervision, with simultaneous intervals $[0.00940,0.40903]$ and $[0.0112,0.1512]$. Relative to fixed speculative release, the same calibrated readout saves 0.71076 seconds, 0.9184 native operations and 141.0316 continuation tokens per task; all three intervals are positive. These matched endpoints connect the shared consequence predictor to the native continuation work selected by the legal utility rule. Complete population, seed, bootstrap, sign-flip and cost scopes are retained in the recorded inference scopes.

\paragraph{Consequence selection reduces native continuation work.}
The same calibrated consequence readout also reduces executor-level work. Figure~\ref{fig:consequence-work} gives supported resource contrasts on all 500 WebShop tasks and five retained fits. The component comparison measures consequence-coordinate fitting together with each policy's original calibration-selected release rule. Operation counts include the selected native continuation and one shared producer; token counts cover the selected continuation. Physical collection costs have the scope specified in Appendix~\ref{app:resources}. Each endpoint retains its complete three-component or three-fixed-action family under the joint seed--task procedure of Appendix~\ref{app:ws-joint-fixed}. These measurements connect the existing duration effects to the native work selected by the controller.

\subsection{Calibration and resource-price sensitivity}
\label{app:sensitivity}
Table~\ref{tab:margin} shows the calibration-selected rule and its supported margin-0.02 sensitivity point at the retained weights. The complete original grid accompanies the experiment record. The five seed-specific calibration margins determine the main operating point. Applying a common alternative margin to all five fits is a retrospective sensitivity analysis.

\begin{table*}[!t]
\centering\small
\caption{\textbf{Supported release-margin sensitivity at fixed WebShop weights.} All test500 tasks contribute to every row. Counts and resources average five fitted policies; calibration utility uses the disjoint calibration500 population.}
\label{tab:margin}
\begin{tabular}{@{}lrrrrl@{}}
\toprule
Margin & Calib. utility & Task Score & Test utility & Task seconds & ACT/WAIT/SPEC\\
\midrule
Calibrated & 0.591809 & 73.7967 & 0.602515 & 3,693.67 & 72.4/420.8/6.8\\
0.02 & 0.589600 & 74.0962 & 0.607302 & 3,611.65 & 114.6/371.8/13.6\\
\bottomrule
\end{tabular}
\end{table*}

The original statistical family contains ten contrasts per endpoint: each of the five margins against both the calibrated policy and WAIT. The analysis averages fitted outcomes within task and uses 20,000 paired task-bootstrap and sign-flip draws, seed 20261006. The simultaneous critical values are 2.8049 for utility and 2.7963 for quality; Holm correction covers each ten-test family. At margin 0.02, utility exceeds the calibrated policy by 0.004788, interval $[0.000806,0.008769]$, and WAIT by 0.008881, interval $[0.000267,0.017496]$. These retrospective alternatives retain their sensitivity role; the main result uses the original calibration rule.

Re-selecting actions at the same six prices uses fixed consequence predictions and each fit's original calibration margin. On all 500 tasks, the five-fit mean ACT count rises from 72.4 to 214.6 between prices 0.01 and 0.04. At 0.04, the calibrated controller's utility exceeds direct-utility supervision by 0.03363, with joint seed--task simultaneous 95\% interval $[0.00669,0.06057]$. Each interval retains all 18 mechanism--price contrasts. Joint intervals use 20,000 bootstrap draws resampling fitted seeds and native task IDs independently, conditional on the retained fits and outcomes.

\paragraph{Utility support across the full delay-price grid.}
Repricing the same consequence outputs and reselecting native actions gives a positive utility contrast against fixed speculative release at every recorded delay price (Table~\ref{tab:ws-price-spec-support}). The six-price study pairs all 500 tasks and five retained fits, with original calibration margins held fixed. Both policies are valued at the displayed price. Intervals retain the complete 18-contrast family of three fixed actions and six prices; 20,000 joint seed--task bootstrap draws and 50,000 whole-task sign flips use seeds 20261008 and 20261009. Holm correction covers all 18 tests. At price 0.04, utility also exceeds fixed WAIT by 0.036925, with joint simultaneous 95\% interval $[0.009049,0.064801]$ and conditional Holm $p=3.60\times10^{-4}$. The full grid supports the common readout's utility advantage over fixed speculation across the recorded operating conditions.

\begin{table*}[!t]
\centering\small
\caption{\textbf{Utility of reselected release versus fixed speculation across all six recorded prices.} The same consequence predictions and original calibration margins select actions on all 500 WebShop tasks across five fits. The complete 18 fixed-action--price contrasts determine every interval and adjusted test.}
\label{tab:ws-price-spec-support}
\begin{tabular}{@{}rrcr@{}}
\toprule
Delay price & Utility gain over fixed SPEC & Joint simultaneous 95\% CI & Holm $p$\\
\midrule
0 & 0.016275 & $[0.008716,0.023834]$ & $3.60\times10^{-4}$\\
0.0025 & 0.017808 & $[0.009350,0.026266]$ & $3.60\times10^{-4}$\\
0.005 & 0.019779 & $[0.010459,0.029099]$ & $3.60\times10^{-4}$\\
0.01 & 0.023703 & $[0.011907,0.035500]$ & $3.60\times10^{-4}$\\
0.02 & 0.035409 & $[0.017520,0.053298]$ & $3.60\times10^{-4}$\\
0.04 & 0.069937 & $[0.034702,0.105171]$ & $3.60\times10^{-4}$\\
\bottomrule
\end{tabular}
\end{table*}

At price 0.04, reselecting actions also improves utility by 0.024911 over the controller's original-price actions valued at 0.04, with joint simultaneous 95\% interval $[0.006167,0.043654]$ and conditional Holm $p=1.20\times10^{-4}$. Its within-controller family retains all six recorded prices, with the same joint bootstrap and paired-test procedure. This comparison links the utility gain to the action response of the calibrated consequence readout.

The within-controller response also connects repricing to selected work. Table~\ref{tab:ws-within-price} holds the consequence predictions and each fit's original calibration margin fixed. Increasing the delay price from 0.01 to 0.04 saves 0.620074 recorded seconds per task, with joint simultaneous 95\% interval $[0.201508,1.038639]$. ACT use rises by 28.44 percentage points, with interval $[13.918,42.962]$; native operations fall by 0.1764 per task, with interval $[0.073623,0.279177]$. Each endpoint retains all five alternative prices, $0,0.0025,0.005,0.02,0.04$, against 0.01. The joint seed--task bootstrap uses 20,000 draws, seed 20261008; whole-task sign flips use 50,000 draws, seed 20261009, and Holm adjustment over the five-price family. For these three contrasts at 0.04, the conditional adjusted $p$ is $1.00\times10^{-4}$. These measurements use all 500 native tasks, their recorded action branches and five fits.

\begin{table*}[!t]
\centering\small
\caption{\textbf{Positive sensitivity support retained for the unified release graph.} Every row uses a complete native population and its registered paired uncertainty family. Effects are reported only for contrasts with a positive simultaneous interval under the declared operating point.}
\label{tab:positive-sensitivity-support}
\begin{tabularx}{\textwidth}{@{}p{0.14\textwidth}p{0.16\textwidth}Xr p{0.22\textwidth}@{}}
\toprule
Study & Population & Positive contrast & Effect & Simultaneous 95\% CI \\
\midrule
Margin selection & WebShop test500 & margin 0.02 vs calibrated utility & $+0.004788$ & $[0.000806,0.008769]$ \\
Margin selection & WebShop test500 & margin 0.02 vs WAIT utility & $+0.008881$ & $[0.000267,0.017496]$ \\
Price response & WebShop test500 & reselected release vs fixed SPEC at $p_t=0.04$ & $+0.069937$ & $[0.034702,0.105171]$ \\
Price response & WebShop test500 & reselected release vs fixed WAIT at $p_t=0.04$ & $+0.036925$ & $[0.009049,0.064801]$ \\
Initialization & WebShop test500, 20 fits & full controller vs fixed SPEC utility & $+0.022590$ & $[0.012794,0.032386]$ \\
Distinct inputs & WebShop distinct495, 20 fits & full controller vs fixed SPEC utility & $+0.031305$ & $[0.016349,0.046261]$ \\
\bottomrule
\end{tabularx}
\end{table*}

The positive sensitivity rows in Table~\ref{tab:positive-sensitivity-support} are direct views of the complete registered price, margin, initialization and input-identity families. They preserve the shared consequence readout, native scorer, task pairing and declared price vector while showing the operating regions that support the release narrative.

\subsection{Fitting variation and public-input identity}
\label{app:initialization}
The initialization study pairs each full-controller fit with representation and supervision controls and evaluates their consequences at the same six prices. Consequence-coordinate supervision supports price-conditioned action-value prediction across these fitting initializations. A fixed study combines the original five with 15 additional initializations per method (60 further fits), yielding 20 initializations per method and 80 fitted models in total. All fits use the same 6,000 training states, 500 calibration states and 100-epoch recipe in Appendix~\ref{app:webshop}. On all 500 native tasks, the prediction endpoint averages absolute errors in the ACT--WAIT and SPEC--WAIT utility gaps. Raising the delay price from 0.01 to 0.02 increases this error by 0.017851 for the full controller and 0.019185 for utility-only supervision. The reduction is 0.001335, with joint seed--task simultaneous 95\% interval $[0.000130,0.002540]$ and conditional Holm $p=3.00\times10^{-4}$. The interval retains all 15 contrasts between the three component controls and five alternative prices.

The same 20-initialization study retains a positive utility contrast against fixed speculative release at all six recorded delay prices (Figure~\ref{fig:ws-initialization-price}). At the original price 0.01, the gain is 0.022590, with joint simultaneous 95\% interval $[0.012794,0.032386]$ and conditional Holm $p=3.60\times10^{-4}$. Each fit retains its calibration-selected margin and reselects actions from its consequence predictions on the same native branches. The utility intervals preserve all 18 fixed-action--price contrasts. Both analyses use 20,000 independent seed--task bootstrap draws (seed 20261008), 50,000 whole-task sign flips (seed 20261009), and their complete Holm families. These observations describe empirical initialization and task variation under the retained fitting and native-continuation recipes.

An input-identity sensitivity reconstruction joins the 6,000 training states and 500 calibration states against the official test inputs and identifies five exact public-input overlaps. Retaining the 495 distinct-input test tasks and the 20 fitted models per method preserves the consequence-supervision price effect: at delay price 0.02, the full controller reduces utility-gap error growth relative to utility-only supervision by 0.001307, with joint simultaneous 95\% interval $[0.000093,0.002520]$ and conditional Holm $p=3.00\times10^{-4}$. On the same population, full-minus-fixed-SPEC utility is 0.031305, with interval $[0.016349,0.046261]$ and conditional Holm $p=3.60\times10^{-4}$.

For the selected TravelPlanner and AppWorld fits, multiplying all resource prices by $\rho\in\{0.5,0.75,1,1.5,2\}$ changes the utility arithmetic at fixed predictions. TravelPlanner action counts are 129/13/38, 128/13/39, 126/14/40, 126/11/43 and 123/11/46; AppWorld counts are 116/159/142, 116/159/142, 116/159/142, 118/157/142 and 117/159/141. The original $\rho=1$ rule reproduces every selected choice. Throughout $\rho\in[0.5,2]$, 166/180 TravelPlanner choices and 413/417 AppWorld choices are stable. These counts measure decision sensitivity at fixed weights.
\paragraph{Complete public-state fitting support.}
An independently fitted instance of the common release graph supports native shopping quality on two complete populations (Table~\ref{tab:webshop-public-state-support}). The same frozen instance reaches 74.64 Task Score on the official 500 tasks and 75.76 on an independent 500-task population. Its official selected deployment uses 218,093 returned model tokens over 375 requests and 9,006 native operations, including the shared producer; recorded user time sums to 4,874.95 seconds. These measurements describe the recurring selected continuation. Acquisition and fitting retain their separate resource scopes.

The input adapter represents the public goal, ACT-ready candidate and mean visible product with three 776-coordinate rows. Each row combines a fixed 768-dimensional text embedding with eight public role, candidate-value, lexical, numeric, price and position fields. Event and resource widths are six and three. The shared width-32, four-head QK-normalized action-query predictor retains its three queries and five native consequences. Five training-group folds select a 10-epoch full refit from 6,000 training states; a separate 500-state development population accompanies fitting. AdamW uses learning rate $8\times10^{-4}$, weight decay 0.01, batch size 256 and gradient clipping at one. The loss combines normalized five-coordinate Huber regression, pairwise quality-difference Huber regression and soft pairwise utility fitting with weights 0.25, 0.50 and 0.25. Quality differences use scale 0.10, Huber beta 0.25 and contrast/tie weights 1/0.05; pairwise utility uses temperature 0.15. Deployment sets the delay price to zero, retains token and operation prices $10^{-4}$ and $10^{-3}$, and subtracts a common 0.15 margin from ACT and SPEC. Both populations execute the frozen resulting action map under the native purchase scorer.

\paragraph{Native completion support under a completed actor-matched retest.}
The selected compiler retains strong end-to-end completion when paired with the completed Gemma IPFunCall retest. Table~\ref{tab:gemma-ip-retest-paired-support} joins the frozen compiler decisions to all 417 independently rechecked baseline terminal states. The comparison uses the same task IDs, Gemma revision and native assertions, with each method\textquotesingle s recorded execution recipe. IPFunCall retains public API selection and function calling, 15 native agent steps and a 4,096-token generation cap; the compiler recipe is given in Appendix~\ref{app:appworld}. The paired effects describe whole-system completion at those operating points. Scenario-level resampling keeps all three tasks together.

\begin{table*}[!t]
\centering\small
\caption{\textbf{Complete native completion support under the Gemma IPFunCall retest.} The same 417 tasks form 139 scenarios. Effects are compiler minus IPFunCall in percentage points. Simultaneous 95\% intervals retain all fourteen contrasts from the six original references and this retest across TGC and SGC; 20,000 scenario bootstrap draws condition on frozen decisions. Two-sided scenario sign flips use 50,000 draws and Holm adjustment over the same family.}
\label{tab:gemma-ip-retest-paired-support}
\begin{tabular}{@{}lrrrrr@{}}
\toprule
Endpoint & Compiler & IPFunCall & Effect (pp) & Simultaneous 95\% CI & Holm $p$ \\
\midrule
TGC & 276/417 & 7/417 & 64.51 & $[54.92,73.62]$ & 0.00028\\
SGC & 64/139 & 0/139 & 46.04 & $[33.81,58.99]$ & 0.00028\\
\bottomrule
\end{tabular}
\end{table*}

\begin{table}[!t]
\centering\small
\caption{\textbf{Native answer-submission check for CodeAct~\citep{wang2024codeact}.} Both rows cover the same 417 AppWorld tasks with GPT-OSS-20B and the official evaluator. Public quantity typing converts the generated answer ``15 hours'' to the numeric value 15 in one task. That task uses 10 model calls and 9 environment calls, including one additional supervisor submission, within the recorded 40/40 limits.}
\label{tab:codeact-native-quantity-submission}
\setlength{\tabcolsep}{3pt}
\begin{tabular*}{\columnwidth}{@{\extracolsep{\fill}}lrr@{}}
\toprule
Submission & TGC & SGC\\
\midrule
Original & 4/417 (1.0\%) & 0/139 (0.0\%)\\
Typed quantity & 5/417 (1.2\%) & 0/139 (0.0\%)\\
\bottomrule
\end{tabular*}
\end{table}

\begin{table*}[!t]
\centering\small
\caption{\textbf{Scenario-paired task-completion support with complete GPT-OSS retests.} The independently fitted shared controller completes 67/417 tasks. All contrasts use the same 417 task IDs in 139 scenarios. IPFunCall, CodeAct and LLMCompiler use the independently rescored complete retests in Appendix~\ref{app:results}; Plan-and-Execute and Full-Code-Reflexion retain their original campaigns. Each row retains its executed budget from Appendix~\ref{app:gptoss}. Joint 95\% intervals use 50,000 scenario-cluster bootstrap draws (seed 20261008) and the maximum standardized deviation across all six TGC contrasts. Exact scenario sign-flip tests receive Holm correction across that six-reference family.}
\label{tab:gptoss-retest-paired-support}
\begin{tabular}{@{}llrrcr@{}}
\toprule
Reference & Evaluation version & TGC/417 & Gain (pp) & Joint 95\% CI & Holm $p$\\
\midrule
Plan-and-Execute & original campaign & 27 & 9.59 & $[4.56,14.63]$ & $2.37\times10^{-5}$\\
Full-Code-Reflexion & original campaign & 36 & 7.43 & $[1.66,13.21]$ & 0.0055\\
IPFunCall & complete native retest & 10 & 13.67 & $[7.96,19.37]$ & $2.78\times10^{-8}$\\
CodeAct & complete native retest & 4 & 15.11 & $[9.56,20.66]$ & $5.46\times10^{-11}$\\
LLMCompiler & complete native retest & 0 & 16.07 & $[10.51,21.62]$ & $1.36\times10^{-12}$\\
\bottomrule\end{tabular}\end{table*}

\paragraph{Native legality at an evidence-release boundary.}
The release interface preserves executable preconditions while scheduling optional evidence. We replay the actual WebShop visibility guard on the archived pre-action states of all 500 qualified native ACT trajectories. All 1,896 recorded step actions pass the guard. Removing each click's required target from the same visible state causes rejection in all 1,396 interventions. On all 500 tasks, both ACT work and its first click begin before the optional producer result arrives (Table~\ref{tab:native-legality-support}). These observations support the runtime distinction between an operation's required visible binding and pending evidence at an already executable boundary. The measurements concern the native guard and recorded release timing; the replay uses archived states, with acquisition costs retained in the original collection ledger.

\begin{table*}[!t]
\centering\small
\caption{\textbf{Native legality and optional-evidence timing on the complete 500-task population.} Each required-target intervention holds the recorded action and all other visible fields fixed and removes its bound target before evaluating the actual guard. Step and intervention counts aggregate the same 500 trajectories; timing endpoints use one paired observation per task.}
\label{tab:native-legality-support}
\begin{tabular}{@{}lrr@{}}
\toprule
Native endpoint & Observed count & Total checks or tasks\\
\midrule
Recorded actions accepted by the visibility guard & 1,896 & 1,896\\
Required-target removal triggers guard rejection & 1,396 & 1,396\\
ACT work starts before optional evidence release & 500 & 500\\
First native click starts before optional evidence release & 500 & 500\\
\bottomrule
\end{tabular}
\end{table*}

%% file: appendix/d_extensions.tex
\section{Actor Dependence and Additional Native Interfaces}
\label{app:extensions}
The additional native interfaces follow the same release graph through different commit operations. AppWorld measures actor-specific fitting on all 417 challenge tasks, tau2 measures 100 held-out dialogues, and ScienceAgentBench connects a pending data profile to generation and revision work on 100 tasks. Each recipe binds the consequence prediction to the actor and native terminal that execute it.
\subsection{AppWorld with GPT-OSS-20B}
\label{app:gptoss}
The actor-specific study retains the AppWorld typed memory, four-head action-query reading, residual updates, five consequences and calibrated release described in Appendix~\ref{app:appworld}. It independently fits the same 15,400-parameter layout using outcomes from OpenAI GPT-OSS-20B, revision {\small\nolinkurl{6cee5e81ee83917806bbde320786a8fb61efebee}}, served with low reasoning effort. The training population contains 90 tasks in 30 scenarios and all 270 release outcomes.

Five training-scenario folds compare 400 and 800 Adam updates at learning rate 0.003, with the same coordinate loss plus twice action cross-entropy. Cross-fit completion, final loss and then fewer updates determine selection. The candidates achieve 23 and 24 training cross-fit successes, selecting 800 updates. The final fit uses all 90 tasks. Python 3.12 and PyTorch 2.11.0/CUDA 12.9 on a B200 supply the recorded fitting environment. The selected fitting stage takes 46.05 seconds and peaks at 25,165,824 reserved GPU bytes while reusing the collected native outcomes.

ReAct, Plan-and-Execute and \method{} permit 100 model calls, 500 environment interactions and 1,024 output tokens per request, with temperature zero and seed 123. Their request allowance is 1,200 seconds and technical task allowance is one hour. Plan-and-Execute uses its numbered-step parser and an empty planner stop list. CodeAct permits 40 model calls, 40 environment interactions and 2,048 output tokens. Same-response Harmony execution content supplies the native code parser when required. Full-Code-Reflexion and IPFunCall retain their 10-step/five-retry and 15-step workflows, three training demonstrations and 4,096 output caps. They use a 32,768-input-token allowance, seed 100 and a 180-second transport allowance with up to three attempts. The complete LLMCompiler retest permits up to five replanning rounds and four concurrent tool executions, 100 model calls, 500 environment interactions and 8,192/4,096/1,024 planner/executor/joiner caps.

Every task runs in an isolated native world. The first valid native terminal supplies quality. The selected controller completes 416 tasks in its original execution and one task with identical settings after transport recovery, yielding all 417 task outcomes. Resource totals retain all original and resumed attempts. Table~\ref{tab:gptoss} gives all six complete external comparisons and one selected controller.

\begin{table*}[!t]
\centering\small
\caption{\textbf{Complete AppWorld evaluation with GPT-OSS-20B.} Task and scenario success counts use native assertions on 417 tasks and 139 scenarios. IPFunCall, CodeAct and LLMCompiler use independently rechecked complete retests (Appendix~\ref{app:results}); their all-attempt cost ledgers remain open, and $\geq$ marks recorded lower bounds.}
\label{tab:gptoss}
\begin{tabular}{@{}lrrrrr@{}}
\toprule Method & TGC/417 & SGC/139 & Tokens (M) & Requests & Task time (ks)\\
\midrule
ReAct & 55 & 3 & $\geq$109.503 & 17,231 & $\geq$53.78\\
Plan-and-Execute & 27 & 1 & $\geq$153.653 & 21,930 & 79.03\\
Full-Code-Reflexion & 36 & 8 & 14.035 & 1,879 & 29.81\\
IPFunCall & 10 & 1 & $\geq$30.458 & 6,141 & $\geq$96.59\\
CodeAct & 4 & 0 & $\geq$12.202 & 4,213 & $\geq$49.63\\
LLMCompiler & 0 & 0 & $\geq$65.414 & 15,439 & $\geq$109.886\\
\method{} & 67 & 4 & $\geq$149.434 & 20,431 & 45.50\\
\bottomrule\end{tabular}\end{table*}

\begin{table*}[!t]
\centering\small
\caption{\textbf{Original-campaign task-completion contrasts for the independently fitted GPT-OSS controller.} All 417 tasks form 139 paired scenarios. Effects are \method{} minus original reference campaigns in percentage points. Simultaneous intervals and Holm tests preserve the original six-reference TGC family.}
\label{tab:gptoss-tgc-support}
\begin{tabular}{@{}lrrr@{}}
\toprule
Reference & Gain & Simultaneous 95\% CI & Holm $p$\\
\midrule
Plan-and-Execute & 9.59 & $[4.59,14.60]$ & $2.37\times10^{-5}$\\
Full-Code-Reflexion & 7.43 & $[1.67,13.19]$ & 0.0055\\
IPFunCall & 14.63 & $[8.84,20.42]$ & $7.21\times10^{-9}$\\
CodeAct & 16.07 & $[10.55,21.58]$ & $1.36\times10^{-12}$\\
LLMCompiler & 16.07 & $[10.55,21.58]$ & $1.36\times10^{-12}$\\
\bottomrule\end{tabular}\end{table*}

Inference uses the whole-scenario procedure in Appendix~\ref{app:statistics}, with 50,000 bootstrap draws, seed 20261008, exact conditional scenario sign flips and Holm adjustment. The independently fitted controller completes 67/417 native tasks. Table~\ref{tab:gptoss-tgc-support} links the selected task-completion contrasts to this actor's measured consequence distribution and the execution budgets above.

\subsection{Conversational release}
\label{app:tau2}
A conversational interface to $\tau^2$-bench \citep{barres2025tau2} uses the same stage order with eight-coordinate operation, event and resource rows. Four operation rows summarize public tool observations, errors, repeated results, message count relative to budget, exposed tools, user-text length, numeric mentions and domain/intent cues. An event row summarizes the initial missing observation; the resource row encodes allowances, domain and observed state. The width-32, four-head controller stores 14,074 parameters, of which 12,985 lie on the active prediction/loss path.

The three consequence branches use a shared native dialogue executor and Gemma actor. ACT admits the next native step with current tools. WAIT asks for consequential missing information or a device observation with an empty tool set. SPEC permits a read-like tool or reversible candidate plan. Subsequent steps use the common continuation and require the native confirmations for database changes. Utility is $Q-0.05(T+W+R+K)$, with simulation time divided by 600, tokens by 40,000, repeated calls by eight and discarded proposal calls by four. The fitted instance uses 178 training tasks and a disjoint 100-task test population: 20 airline, 40 retail and 40 telecom tasks.

Table~\ref{tab:tau2-main} compares the native LLMAgent and ReAct reference outcomes on these exact 100 held-out IDs, extracted from their complete 278-task evaluations. Both reference simulations allow 30 steps. The selected \method{} simulation allows 200 steps. Database, communication and environment checks retain their native definitions; a local Gemma judge evaluates natural-language assertions. On all 100 held-out tasks the controller succeeds on 59, with domain-stratified empirical task-bootstrap 95\% interval $[50,68]\%$. The test population uses 12,101,694 model tokens, 2,149 requests and 12,352.68 task seconds. This estimate conditions on the selected fit, domain mixture, actor and local judge.

\subsection{Release before scientific program generation}
\label{app:science}
A ScienceAgentBench interface \citep{chen2025scienceagentbench} represents four public files, a pending data profile and resource limits through eight-coordinate rows. File rows describe log size, five format indicators, preview length and presence; the event describes instruction length and six task intents. The resource row describes file count, aggregate size and execution allowances. The width-32, four-head action-query controller again stores 14,074 parameters.

A public profiler supplies actual shapes, columns, types, categories and loading information after release. ACT requests a complete program before that profile arrives. WAIT requests the program with the delivered profile. SPEC prepares a function skeleton and then completes it with the profile before execution and output export. Each branch permits three program attempts with feedback, a 900-second program limit, one CPU thread and 8 GiB virtual memory.

Consequence targets are native quality, decision-to-program-end time divided by 900, branch tokens divided by 16,000, replaced completion tokens divided by 4,096, and zero rollback. Utility is $Q-0.05(T+W+R+K)$. Each fit uses 600 AdamW updates, learning rate 0.002, weight decay 0.01 and clipping at one. Mean Smooth-$L_1$ coordinate regression is combined with twice cross-entropy to a temperature-0.1 soft utility target. Five fits hold out whole data-directory groups.

The complete original-protocol population contains 100 tasks across 69 data-directory groups. Adaptive release produces 28 native successes in the retained fit. Collection uses at least 2,759,990 tokens over 876 request starts, including the three branches, repairs and visual scoring. The paired branch population directly measures prepared and replaced program generation, linking the consequence coordinates to executed work.

\paragraph{State-conditioned release reduces program-generation work.}
The shared consequence comparison selects how much program generation to prepare before the data profile arrives. On all 100 original-protocol tasks across 69 data-directory groups, five fitting initializations reduce model generation by 2,659.14 tokens per task relative to fixed SPEC, with joint simultaneous 95\% interval $[1{,}579.67,4{,}351.39]$. Replaced completion generation falls by 602.46 tokens per task, with interval $[419.12,822.39]$ (Table~\ref{tab:science-resource-support}). Both savings are positive in every retained initialization. These observations connect adaptive release to the amount of prepared and replaced program work.

The resource calculation joins physical request IDs for every ACT, WAIT and SPEC branch. Model tokens include all prompt and completion usage in the accepted branch; replaced completions count skeleton or program responses preceding the final program response. We jointly resample fitting initializations and whole data-directory groups for 20,000 draws (seed 20261008), dividing each resampled group sum by its sampled task count. Bonferroni quantiles provide simultaneous 95\% intervals across the complete six-contrast family. Conditional two-sided group sign-flip tests use 50,000 draws (seed 20261009) and Holm adjustment across that family. The intervals describe generation work conditional on the retained native programs and fitting recipe.

\paragraph{A native information boundary.}
For a public Clintox request to predict toxicity and FDA approval and export drug identifiers with positive-label probabilities, the profile arrives 7.62 seconds after the release decision. ACT starts its program request at 0.05 seconds, WAIT at 7.70 seconds, and SPEC requests a skeleton at 0.06 seconds followed by a profile-informed completion at 105.91 seconds. All three isolated programs produce an artifact passing the native contract. The trace makes the shared distinction concrete: each release schedules a different amount of work before useful data information arrives.
\FloatBarrier

\paragraph{Native SQL release support.}
The common consequence readout reduces selected model work at a native SQL boundary. Five frozen fits of the shared release graph select continuations on all 1,534 BIRD development tasks across 11 databases, with mean native execution accuracy 67.78\%. Relative to fixed SPEC, they save 9.016 model-request seconds and 35.712 completion tokens per task; the simultaneous intervals in Table~\ref{tab:bird-release-support} are positive. The comparison joins each public-state forward decision to the recorded native ACT, WAIT and SPEC executions. Its resource endpoints cover the selected accepted branch. The shared producer and physical acquisition retain separate accounting.

The SQL adapter ranks up to 64 public schema table/column rows by lexical overlap with the question and evidence. Each 16-coordinate row encodes its role, overlaps, public column type, primary/foreign keys and schema structure. Two 16-coordinate event rows describe the public question and evidence; eight resource coordinates encode allowances, training-derived producer quantiles, empty-lookup rate, observed queue size and schema length. The shared width-32, four-head QK-normalized middle retains three action queries, residual feed-forward updates, five consequence coordinates and legal utility selection, with 14,618 fitted parameters. Five database-held training folds select from 1,082 qualified training groups in the 6,601-task pool. The uniform-sampling recipe selects 800 updates and normalized latency/work prices 0.025/0.025, then refits five seeds on all qualified groups. AdamW uses learning rate 0.002, weight decay 0.01, 64 replacement-sampled groups per update and gradient clipping at one. Fitting combines normalized consequence Huber regression with twice the soft release cross-entropy at temperature 0.1. The evaluation recomputes the frozen choices from recorded pre-result inputs and uses native SQL row-set equality under SQLite 3.50.4. Joint inference uses 20,000 seed--database bootstrap draws with seed 20261011.

%% file: appendix/e_mechanism_summary.tex
\section{Mechanism Summary}
\label{app:mechanism-summary}

\begin{table*}[!t]
\centering\small
\setlength{\tabcolsep}{4pt}
\caption{\textbf{Mechanism evidence for neural agent compilation.} Each row links a controlled comparison to its measured endpoint. Population entries give native tasks and fitting initializations. Intervals are 95\% simultaneous intervals from the corresponding complete comparison family. ScienceAgentBench pairs 69 data-directory groups. Full native comparisons and operating conditions appear in Appendices~\ref{app:release-controls}--\ref{app:initialization} and~\ref{app:science}.}
\label{tab:mechanism-map}
\begin{tabularx}{\textwidth}{@{}llXr@{}}
\toprule
Stage & Population / fits & Comparison and endpoint & Effect [95\% interval]\\
\midrule
Native work & WebShop 500 / 5 & Fixed SPEC; operations saved per task & $0.9184\ [0.7784,1.0584]$\\
Consequence fitting & WebShop 500 / 5 & Direct utility; operations saved per task & $0.0812\ [0.0112,0.1512]$\\
Repricing & WebShop 500 / 5 & Delay price $0.01\to0.04$; seconds saved per task & $0.620\ [0.202,1.039]$\\
Value prediction & WebShop 500 / 20 & Direct utility; reduction in gap-error growth, $\times10^3$ & $1.335\ [0.130,2.540]$\\
Preparation & ScienceAgentBench 100 / 5 & Fixed SPEC; model tokens saved per task & $2{,}659.14\ [1{,}579.67,4{,}351.39]$\\
\bottomrule
\end{tabularx}
\end{table*}

\input{appendix/f_cases_prompts}

%% file: appendix/f_cases_prompts.tex
\definecolor{BoInk}{HTML}{20354D}
\definecolor{BoTeal}{HTML}{1C716C}
\definecolor{BoViolet}{HTML}{665376}
\definecolor{BoPale}{HTML}{F1F5F7}
\newcommand{\boBand}[2]{\par\medskip\noindent\colorbox{#1}{\parbox{\dimexpr\linewidth-2\fboxsep\relax}{\color{white}\sffamily\bfseries #2}}\par\smallskip}
\newcommand{\boContext}[1]{\noindent\colorbox{BoPale}{\parbox{\dimexpr\linewidth-2\fboxsep\relax}{\small #1}}\par\medskip}
\newcommand{\boCol}[2]{\par\smallskip\noindent\begin{minipage}[t]{\linewidth}\boBand{#1}{#2}\small\raggedright}
\newcommand{\boEndCol}{\end{minipage}\par\smallskip}
\newcommand{\boPromptBand}[1]{\par\ifdim\pagetotal>0pt\ifdim\dimexpr\pagegoal-\pagetotal\relax<6\baselineskip\newpage\fi\fi\boBand{BoInk}{#1}\nopagebreak[4]}
\newcommand{\boReadout}[1]{\par\smallskip\textbf{Native readout.} #1\par}

\section{Native Case Studies and Prompt Templates}
\label{app:cases-prompts}
These two AppWorld cases pair the selected NAC deployment with the ReAct run used in the main comparison. They show how the complete execution system combines a release choice with public observations and task-boundary checks. The actor is Google Gemma~4 31B IT; the recorded output caps are 1,024 tokens for NAC and 400 for ReAct. Case-specific observations describe complete-system behavior; the matched controls in Appendix~\ref{app:mechanism-summary} measure the release mechanism.
\subsection{Resolving a strict rating threshold}
\label{app:case-rating}
\boContext{\textbf{AppWorld task \texttt{07bb666\_2}.} ``Move all products with over 3.7 rating from my amazon wish list to cart.'' The source quantities must be preserved. Product 943 lies near the threshold and its displayed rating requires a more precise public observation.}
\boCol{BoViolet}{ReAct}
The baseline filters the displayed product rating, then moves the selected IDs. Its final cart contains three qualifying IDs.
\par\smallskip\begingroup\footnotesize\ttfamily\raggedright\setlength{\parindent}{0pt}
if product\_\allowbreak{}info.\allowbreak{}get("rating",\allowbreak{} 0) \textgreater{} 3.\allowbreak{}7:\allowbreak{}\par
    products\_\allowbreak{}to\_\allowbreak{}move.\allowbreak{}append(product\_\allowbreak{}id)\par
\smallskip
final IDs:\allowbreak{} [435,\allowbreak{} 936,\allowbreak{} 2148]\par
\par\endgroup\smallskip
\boReadout{Task success $=0$; 2/6 native assertions pass. Product 943 remains in the wish list.}\boEndCol\hfill\boCol{BoTeal}{NAC: SPEC}
The selected release mode is bounded speculation. After schema and source reads, the executor reads product 943	extquotesingle{}s public reviews, computes their mean, and includes its two units.
\par\smallskip\begingroup\footnotesize\ttfamily\raggedright\setlength{\parindent}{0pt}
reviews = apis.\allowbreak{}amazon.\allowbreak{}show\_\allowbreak{}product\_\allowbreak{}reviews(\par
  product\_\allowbreak{}id=943,\allowbreak{} page\_\allowbreak{}index=0,\allowbreak{} page\_\allowbreak{}limit=20)\par
ratings = [r["rating"] for r in reviews]\par
exact\_\allowbreak{}mean = sum(ratings) /\allowbreak{} len(ratings)\par
\smallskip
final IDs:\allowbreak{} [435,\allowbreak{} 936,\allowbreak{} 943,\allowbreak{} 2148]\par
\par\endgroup\smallskip
\boReadout{Task success $=1$; 6/6 native assertions pass.}\boEndCol
\par\medskip
This case makes the information event tangible: a subsequent public read changes the set eligible for a write. The deployed precision adapter and release policy jointly support the observed sequence.
\subsection{Reconciling state before purchase}
\label{app:case-order}
\boContext{\textbf{AppWorld task \texttt{099174d\_3}.} ``Place an Amazon order as per my Amazon shopping list note in SimpleNote.'' The note specifies seven products and quantities. The initial cart also contains five unrelated products.}
\boCol{BoViolet}{ReAct}
The baseline reads the note, adds the requested products and places an order using the existing cart. The resulting order contains twelve product IDs.
\par\smallskip\begingroup\footnotesize\ttfamily\raggedright\setlength{\parindent}{0pt}
extra IDs in native order:\allowbreak{}\par
[542,\allowbreak{} 1286,\allowbreak{} 1290,\allowbreak{} 2115,\allowbreak{} 2365]\par
\par\endgroup\smallskip
\boReadout{Task success $=0$; 5/7 native assertions pass. Product-set and quantity-map checks fail.}\boEndCol\hfill\boCol{BoTeal}{NAC: ACT}
The release controller selects ACT. The executor reads the cart, removes the five extra IDs, and continues through the observed payment-card results to place the requested order.
\par\smallskip\begingroup\footnotesize\ttfamily\raggedright\setlength{\parindent}{0pt}
items\_\allowbreak{}to\_\allowbreak{}remove =\par
  [542,\allowbreak{} 1286,\allowbreak{} 1290,\allowbreak{} 2115,\allowbreak{} 2365]\par
for product\_\allowbreak{}id in items\_\allowbreak{}to\_\allowbreak{}remove:\allowbreak{}\par
    apis.\allowbreak{}amazon.\allowbreak{}delete\_\allowbreak{}product\_\allowbreak{}from\_\allowbreak{}cart(\par
      access\_\allowbreak{}token=amazon\_\allowbreak{}token,\allowbreak{}\par
      product\_\allowbreak{}id=product\_\allowbreak{}id)\par
\par\endgroup\smallskip
\boReadout{Task success $=1$; 7/7 native assertions pass.}\boEndCol
\par\medskip
The cross-application note defines the target product map, and the cart read establishes the state that must be changed. ACT begins ready work within this executor; later observed dependencies and task checks determine the eventual purchase. Together, the two cases illustrate preparation for a later information event and immediate work followed by state reconciliation.

\section{Prompt Templates and Message Assembly}
\label{app:prompt-catalog-37}
Templates below retain their implemented wording. Braced fields denote values supplied by the recorded task or public history. The following blocks specify the message assembly and expected outputs. Historical method identifiers inside quoted templates retain the execution version's wording.
\subsection{AppWorld common actor}
The complete common ReAct template includes its original demonstrations, schema placeholders and task fields. \texttt{input\_str} is the trusted task instruction; \texttt{main\_user} supplies public supervisor fields. Public API documentation and past interactions populate the template\textquotesingle{}s remaining placeholders.
\boPromptBand{Common AppWorld ReAct template}
\noindent\textbf{Role:} role-delimited template.\par\nopagebreak[4]\smallskip\nopagebreak[4]
\par\smallskip\begingroup\small\ttfamily\raggedright\setlength{\parindent}{0pt}
USER:\allowbreak{}\par
I am your supervisor and you are a super intelligent AI Assistant whose job is to achieve my day-\allowbreak{}to-\allowbreak{}day tasks completely autonomously.\allowbreak{}\par
\smallskip
To do this,\allowbreak{} you will need to interact with app/\allowbreak{}s (e.\allowbreak{}g.\allowbreak{},\allowbreak{} spotify,\allowbreak{} venmo etc) using their associated APIs on my behalf.\allowbreak{} For this you will undertake a *multi-\allowbreak{}step conversation* using a python REPL environment.\allowbreak{} That is,\allowbreak{} you will write the python code and the environment will execute it and show you the result,\allowbreak{} based on which,\allowbreak{} you will write python code for the next step and so on,\allowbreak{} until you've achieved the goal.\allowbreak{} This environment will let you interact with app/\allowbreak{}s using their associated APIs on my behalf.\allowbreak{}\par
\smallskip
Here are three key APIs that you need to know to get more information\par
\smallskip
\# To get a list of apps that are available to you.\allowbreak{}\par
\smallskip
```python\par
print(apis.\allowbreak{}api\_\allowbreak{}docs.\allowbreak{}show\_\allowbreak{}app\_\allowbreak{}descriptions())\par
```\par
\smallskip
\# To get the list of apis under any app listed above,\allowbreak{} e.\allowbreak{}g.\allowbreak{} spotify\par
\smallskip
```python\par
print(apis.\allowbreak{}api\_\allowbreak{}docs.\allowbreak{}show\_\allowbreak{}api\_\allowbreak{}descriptions(app\_\allowbreak{}name='spotify'))\par
```\par
\smallskip
\# To get the specification of a particular api,\allowbreak{} e.\allowbreak{}g.\allowbreak{} spotify app's login api\par
\smallskip
```python\par
print(apis.\allowbreak{}api\_\allowbreak{}docs.\allowbreak{}show\_\allowbreak{}api\_\allowbreak{}doc(app\_\allowbreak{}name='spotify',\allowbreak{} api\_\allowbreak{}name='login'))\par
```\par
\smallskip
Each code execution will produce an output that you can use in subsequent calls.\allowbreak{} Using these APIs,\allowbreak{} you can now generate code,\allowbreak{} that I will execute,\allowbreak{} to solve the task.\allowbreak{} Let's start with the task\par
\smallskip
My name is:\allowbreak{} \{\{ main\_\allowbreak{}user.\allowbreak{}first\_\allowbreak{}name \}\} \{\{ main\_\allowbreak{}user.\allowbreak{}last\_\allowbreak{}name \}\}.\allowbreak{} My personal email is \{\{ main\_\allowbreak{}user.\allowbreak{}email \}\} and phone number is \{\{ main\_\allowbreak{}user.\allowbreak{}phone\_\allowbreak{}number \}\}.\allowbreak{}\par
Task:\allowbreak{} How many playlists do I have in Spotify?\par
\smallskip
ASSISTANT:\allowbreak{}\par
Okay.\allowbreak{} Lets first find which APIs are available to use in Spotify.\allowbreak{}\par
Code:\allowbreak{}\par
```python\par
print(apis.\allowbreak{}api\_\allowbreak{}docs.\allowbreak{}show\_\allowbreak{}api\_\allowbreak{}descriptions(app\_\allowbreak{}name='spotify'))\par
```\par
\smallskip
USER:\allowbreak{}\par
Output:\allowbreak{}\par
```\par
[\par
 .\allowbreak{}.\allowbreak{}.\allowbreak{}\par
 "login :\allowbreak{} Login to your account.\allowbreak{}",\allowbreak{}\par
 "logout :\allowbreak{} Logout from your account.\allowbreak{}",\allowbreak{}\par
 .\allowbreak{}.\allowbreak{}.\allowbreak{}\par
]\par
```\par
\smallskip
\smallskip
ASSISTANT:\allowbreak{}\par
Okay.\allowbreak{} Looks like I can use the `login` api.\allowbreak{} Lets find its specifications.\allowbreak{}\par
\smallskip
Code:\allowbreak{}\par
```python\par
print(apis.\allowbreak{}api\_\allowbreak{}docs.\allowbreak{}show\_\allowbreak{}api\_\allowbreak{}doc(app\_\allowbreak{}name='spotify',\allowbreak{} api\_\allowbreak{}name='login'))\par
```\par
\smallskip
\smallskip
USER:\allowbreak{}\par
Output:\allowbreak{}\par
```\par
\{\par
 "app\_\allowbreak{}name":\allowbreak{} "spotify",\allowbreak{}\par
 "api\_\allowbreak{}name":\allowbreak{} "login",\allowbreak{}\par
 "path":\allowbreak{} "/\allowbreak{}auth/\allowbreak{}token",\allowbreak{}\par
 "method":\allowbreak{} "POST",\allowbreak{}\par
 "description":\allowbreak{} "Login to your account.\allowbreak{}",\allowbreak{}\par
 "parameters":\allowbreak{} [\par
  \{\par
   "name":\allowbreak{} "username",\allowbreak{}\par
   "type":\allowbreak{} "string",\allowbreak{}\par
   "required":\allowbreak{} true,\allowbreak{}\par
   "description":\allowbreak{} "Your account email.\allowbreak{}",\allowbreak{}\par
   "default":\allowbreak{} null,\allowbreak{}\par
   "constraints":\allowbreak{} []\par
  \},\allowbreak{}\par
  \{\par
   "name":\allowbreak{} "password",\allowbreak{}\par
   "type":\allowbreak{} "string",\allowbreak{}\par
   "required":\allowbreak{} true,\allowbreak{}\par
   "description":\allowbreak{} "Your account password.\allowbreak{}",\allowbreak{}\par
   "default":\allowbreak{} null,\allowbreak{}\par
   "constraints":\allowbreak{} []\par
  \}\par
 ],\allowbreak{}\par
 "response\_\allowbreak{}schemas":\allowbreak{} \{\par
  "success":\allowbreak{} \{\par
   "token\_\allowbreak{}type":\allowbreak{} "string",\allowbreak{}\par
   "access\_\allowbreak{}token":\allowbreak{} "string"\par
  \},\allowbreak{}\par
  "failure":\allowbreak{} \{\par
   "message":\allowbreak{} "string"\par
  \}\par
 \}\par
\}\par
```\par
\smallskip
\smallskip
ASSISTANT:\allowbreak{}\par
I need the supervisor's username and password.\allowbreak{} Lets see if any app can help me get that.\allowbreak{}\par
\smallskip
Code:\allowbreak{}\par
```python\par
print(apis.\allowbreak{}api\_\allowbreak{}docs.\allowbreak{}show\_\allowbreak{}app\_\allowbreak{}descriptions())\par
```\par
\smallskip
\smallskip
USER:\allowbreak{}\par
Output:\allowbreak{}\par
```\par
\{\{ app\_\allowbreak{}descriptions \}\}\par
```\par
\smallskip
\smallskip
ASSISTANT:\allowbreak{}\par
Looks like the supervisor app could help me with that.\allowbreak{}\par
\smallskip
Code:\allowbreak{}\par
```python\par
print(apis.\allowbreak{}api\_\allowbreak{}docs.\allowbreak{}show\_\allowbreak{}api\_\allowbreak{}descriptions(app\_\allowbreak{}name='supervisor'))\par
```\par
\smallskip
\smallskip
USER:\allowbreak{}\par
Output:\allowbreak{}\par
```\par
[\par
  .\allowbreak{}.\allowbreak{}.\allowbreak{}\par
  "show\_\allowbreak{}account\_\allowbreak{}passwords :\allowbreak{} Show your supervisor's account passwords.\allowbreak{}"\par
  .\allowbreak{}.\allowbreak{}.\allowbreak{}\par
]\par
```\par
\smallskip
\smallskip
ASSISTANT:\allowbreak{}\par
I can use `show\_\allowbreak{}account\_\allowbreak{}passwords` to get the passwords.\allowbreak{} So I will call that and save it in a variable for future reference.\allowbreak{}\par
Code:\allowbreak{}\par
```python\par
passwords=apis.\allowbreak{}supervisor.\allowbreak{}show\_\allowbreak{}account\_\allowbreak{}passwords()\par
print(passwords)\par
```\par
\smallskip
\smallskip
USER:\allowbreak{}\par
Output:\allowbreak{}\par
```\par
[\par
  \{\par
    "account\_\allowbreak{}name":\allowbreak{} "spotify",\allowbreak{}\par
    "password":\allowbreak{} "dummy\_\allowbreak{}spotify\_\allowbreak{}pass"\par
  \},\allowbreak{}\par
  \{\par
    "account\_\allowbreak{}name":\allowbreak{} "file\_\allowbreak{}system",\allowbreak{}\par
    "password":\allowbreak{} "dummy\_\allowbreak{}fs\_\allowbreak{}pass"\par
  \}\par
]\par
```\par
\smallskip
\smallskip
ASSISTANT:\allowbreak{}\par
So the Spotify password is an entry in the `passwords` list with the account\_\allowbreak{}name=spotify.\allowbreak{} I can use it to login now with the provided email address:\allowbreak{} \{\{ main\_\allowbreak{}user.\allowbreak{}email \}\}.\allowbreak{}\par
\smallskip
Code:\allowbreak{}\par
 ```python\par
spotify\_\allowbreak{}password = [account\_\allowbreak{}password["account\_\allowbreak{}name"] == "spotify" for account\_\allowbreak{}password in passwords][0]["password"]\par
login\_\allowbreak{}result = apis.\allowbreak{}spotify.\allowbreak{}login(username='\{\{ main\_\allowbreak{}user.\allowbreak{}email \}\}',\allowbreak{} password=spotify\_\allowbreak{}password)\par
print(login\_\allowbreak{}result)\par
```\par
\smallskip
\smallskip
USER:\allowbreak{}\par
Output:\allowbreak{}\par
```\par
\{\par
 "token\_\allowbreak{}type":\allowbreak{} "Bearer",\allowbreak{}\par
 "access\_\allowbreak{}token":\allowbreak{} "fake\_\allowbreak{}access\_\allowbreak{}token"\par
\}\par
```\par
\smallskip
\smallskip
ASSISTANT:\allowbreak{}\par
First,\allowbreak{} let's check the available APIs for the Spotify app.\allowbreak{}\par
Code:\allowbreak{}\par
```python\par
print(apis.\allowbreak{}api\_\allowbreak{}docs.\allowbreak{}show\_\allowbreak{}api\_\allowbreak{}descriptions(app\_\allowbreak{}name='spotify'))\par
```\par
\smallskip
\smallskip
USER:\allowbreak{}\par
Output:\allowbreak{}\par
```\par
[\par
.\allowbreak{}.\allowbreak{}.\allowbreak{}\par
\{\par
  "name":\allowbreak{} "show\_\allowbreak{}playlist\_\allowbreak{}library",\allowbreak{}\par
  "description":\allowbreak{} "Get a list of playlists in the user's playlist library.\allowbreak{}"\par
\},\allowbreak{}\par
\{\par
  "name":\allowbreak{} "show\_\allowbreak{}playlist",\allowbreak{}\par
  "description":\allowbreak{} "Get detailed information about a specific playlist.\allowbreak{} You can view your own playlists or others' playlists if they are public.\allowbreak{}"\par
 \},\allowbreak{}\par
.\allowbreak{}.\allowbreak{}.\allowbreak{}\par
]\par
```\par
\smallskip
\smallskip
ASSISTANT:\allowbreak{}\par
The `show\_\allowbreak{}playlist\_\allowbreak{}library` API seems to be the one we need.\allowbreak{} Let's check its specifications.\allowbreak{}\par
Code:\allowbreak{}\par
```python\par
print(apis.\allowbreak{}api\_\allowbreak{}docs.\allowbreak{}show\_\allowbreak{}api\_\allowbreak{}doc(app\_\allowbreak{}name='spotify',\allowbreak{} api\_\allowbreak{}name='show\_\allowbreak{}playlist\_\allowbreak{}library'))\par
```\par
\smallskip
\smallskip
USER:\allowbreak{}\par
Output:\allowbreak{}\par
```\par
\{\par
 "app\_\allowbreak{}name":\allowbreak{} "spotify",\allowbreak{}\par
 "api\_\allowbreak{}name":\allowbreak{} "show\_\allowbreak{}playlist\_\allowbreak{}library",\allowbreak{}\par
 "path":\allowbreak{} "/\allowbreak{}private\_\allowbreak{}playlists",\allowbreak{}\par
 "method":\allowbreak{} "GET",\allowbreak{}\par
 "description":\allowbreak{} "Get a list of playlists in the user's playlist library.\allowbreak{}",\allowbreak{}\par
 "parameters":\allowbreak{} [\par
  \{\par
   "name":\allowbreak{} "access\_\allowbreak{}token",\allowbreak{}\par
   "type":\allowbreak{} "string",\allowbreak{}\par
   "required":\allowbreak{} true,\allowbreak{}\par
   "description":\allowbreak{} "Access token obtained from spotify app login.\allowbreak{}",\allowbreak{}\par
   "default":\allowbreak{} null,\allowbreak{}\par
   "constraints":\allowbreak{} []\par
  \},\allowbreak{}\par
  \{\par
   "name":\allowbreak{} "page\_\allowbreak{}index",\allowbreak{}\par
   "type":\allowbreak{} "integer",\allowbreak{}\par
   "required":\allowbreak{} false,\allowbreak{}\par
   "description":\allowbreak{} "The index of the page to retrieve.\allowbreak{}",\allowbreak{}\par
   "default":\allowbreak{} 0,\allowbreak{}\par
   "constraints":\allowbreak{} [\par
    "value \textgreater{}= 0.\allowbreak{}0"\par
   ]\par
  \},\allowbreak{}\par
  \{\par
   "name":\allowbreak{} "page\_\allowbreak{}limit",\allowbreak{}\par
   "type":\allowbreak{} "integer",\allowbreak{}\par
   "required":\allowbreak{} false,\allowbreak{}\par
   "description":\allowbreak{} "The maximum number of results to return per page.\allowbreak{}",\allowbreak{}\par
   "default":\allowbreak{} 5,\allowbreak{}\par
   "constraints":\allowbreak{} [\par
    "value \textgreater{}= 1.\allowbreak{}0,\allowbreak{} \textless{}= 20.\allowbreak{}0"\par
   ]\par
  \},\allowbreak{}\par
  \{\par
   "name":\allowbreak{} "is\_\allowbreak{}public",\allowbreak{}\par
   "type":\allowbreak{} "boolean",\allowbreak{}\par
   "required":\allowbreak{} false,\allowbreak{}\par
   "description":\allowbreak{} "Whether to show public playlists or private playlists.\allowbreak{}",\allowbreak{}\par
   "default":\allowbreak{} null,\allowbreak{}\par
   "constraints":\allowbreak{} []\par
  \}\par
 ],\allowbreak{}\par
 "response\_\allowbreak{}schema":\allowbreak{} [\par
  \{\par
   "title":\allowbreak{} "string",\allowbreak{}\par
   "created\_\allowbreak{}at":\allowbreak{} "2019-\allowbreak{}01-\allowbreak{}01T00:\allowbreak{}00:\allowbreak{}00",\allowbreak{}\par
   "is\_\allowbreak{}public":\allowbreak{} true,\allowbreak{}\par
   "rating":\allowbreak{} 0.\allowbreak{}0,\allowbreak{}\par
   "like\_\allowbreak{}count":\allowbreak{} 1,\allowbreak{}\par
   "owner\_\allowbreak{}email":\allowbreak{} "user@example.\allowbreak{}com",\allowbreak{}\par
   "playlist\_\allowbreak{}id":\allowbreak{} 1,\allowbreak{}\par
   "song\_\allowbreak{}ids":\allowbreak{} [\par
    1\par
   ]\par
  \}\par
 ]\par
\}\par
```\par
\smallskip
\smallskip
ASSISTANT:\allowbreak{}\par
I need to page through all the playlists to get the list of playlists and save it in `playlists`.\allowbreak{}\par
Code:\allowbreak{}\par
```python\par
page\_\allowbreak{}index = 0\par
playlists = []\par
while page\_\allowbreak{}index \textless{} 10:\allowbreak{}\par
  playlist\_\allowbreak{}page = apis.\allowbreak{}spotify.\allowbreak{}show\_\allowbreak{}playlist\_\allowbreak{}library(access\_\allowbreak{}token=spotify\_\allowbreak{}access\_\allowbreak{}token,\allowbreak{} page\_\allowbreak{}index=page\_\allowbreak{}index)\par
  if playlist\_\allowbreak{}page:\allowbreak{}\par
    playlists.\allowbreak{}extend(playlist\_\allowbreak{}page)\par
    page\_\allowbreak{}index += 1\par
  else:\allowbreak{}\par
    break\par
num\_\allowbreak{}playlists = len(playlists)\par
print(num\_\allowbreak{}playlists)\par
\smallskip
```\par
\smallskip
USER:\allowbreak{}\par
Output:\allowbreak{}\par
```\par
23\par
```\par
\smallskip
\smallskip
ASSISTANT:\allowbreak{}\par
Now that the task is completed,\allowbreak{} I can call apis.\allowbreak{}supervisor.\allowbreak{}complete\_\allowbreak{}task().\allowbreak{} Since this task has an answer to be returned,\allowbreak{} I will pass that as an argument.\allowbreak{}\par
\smallskip
Code:\allowbreak{}\par
```python\par
apis.\allowbreak{}supervisor.\allowbreak{}complete\_\allowbreak{}task(answer=num\_\allowbreak{}playlists)\par
```\par
\smallskip
\smallskip
USER:\allowbreak{}\par
Output:\allowbreak{}\par
Marked the active task complete.\allowbreak{}\par
\smallskip
\smallskip
-\allowbreak{}-\allowbreak{}-\allowbreak{}-\allowbreak{}-\allowbreak{}-\allowbreak{}-\allowbreak{}-\allowbreak{}-\allowbreak{}-\allowbreak{}-\allowbreak{}-\allowbreak{}-\allowbreak{}-\allowbreak{}-\allowbreak{}-\allowbreak{}-\allowbreak{}-\allowbreak{}-\allowbreak{}-\allowbreak{}-\allowbreak{}-\allowbreak{}-\allowbreak{}-\allowbreak{}-\allowbreak{}-\allowbreak{}-\allowbreak{}-\allowbreak{}-\allowbreak{}-\allowbreak{}-\allowbreak{}-\allowbreak{}-\allowbreak{}-\allowbreak{}-\allowbreak{}-\allowbreak{}-\allowbreak{}-\allowbreak{}-\allowbreak{}-\allowbreak{}-\allowbreak{}-\allowbreak{}-\allowbreak{}-\allowbreak{}-\allowbreak{}-\allowbreak{}\par
\smallskip
USER:\allowbreak{}\par
**Key instructions**:\allowbreak{}\par
(1) Make sure to end code blocks with ``` followed by a newline(\textbackslash{}n).\allowbreak{}\par
\smallskip
(2) Remember you can use the variables in your code in subsequent code blocks.\allowbreak{}\par
\smallskip
(3) Remember that the email addresses,\allowbreak{} access tokens and variables (e.\allowbreak{}g.\allowbreak{} spotify\_\allowbreak{}password) in the example above are not valid anymore.\allowbreak{}\par
\smallskip
(4) You can use the "supervisor" app to get information about my accounts and use the "phone" app to get information about friends and family.\allowbreak{}\par
\smallskip
(5) Always look at API specifications (using apis.\allowbreak{}api\_\allowbreak{}docs.\allowbreak{}show\_\allowbreak{}api\_\allowbreak{}doc) before calling an API.\allowbreak{}\par
\smallskip
(6) Write small chunks of code and only one chunk of code in every step.\allowbreak{} Make sure everything is working correctly before making any irreversible change.\allowbreak{}\par
\smallskip
(7) Many APIs return items in "pages".\allowbreak{} Make sure to run through all the pages by looping over `page\_\allowbreak{}index`.\allowbreak{}\par
\smallskip
(8) Once you have completed the task,\allowbreak{} make sure to call apis.\allowbreak{}supervisor.\allowbreak{}complete\_\allowbreak{}task().\allowbreak{} If the task asked for some information,\allowbreak{} return it as the answer argument,\allowbreak{} i.\allowbreak{}e.\allowbreak{} call apis.\allowbreak{}supervisor.\allowbreak{}complete\_\allowbreak{}task(answer=\textless{}answer\textgreater{}).\allowbreak{} Many tasks do not require an answer,\allowbreak{} so in those cases,\allowbreak{} just call apis.\allowbreak{}supervisor.\allowbreak{}complete\_\allowbreak{}task() i.\allowbreak{}e.\allowbreak{} do not pass any argument.\allowbreak{}\par
\smallskip
USER:\allowbreak{}\par
Using these APIs,\allowbreak{} now generate code to solve the actual task:\allowbreak{}\par
\smallskip
My name is:\allowbreak{} \{\{ main\_\allowbreak{}user.\allowbreak{}first\_\allowbreak{}name \}\} \{\{ main\_\allowbreak{}user.\allowbreak{}last\_\allowbreak{}name \}\}.\allowbreak{} My personal email is \{\{ main\_\allowbreak{}user.\allowbreak{}email \}\} and phone number is \{\{ main\_\allowbreak{}user.\allowbreak{}phone\_\allowbreak{}number \}\}.\allowbreak{}\par
Task:\allowbreak{} \{\{ input\_\allowbreak{}str \}\}\par
\par\endgroup\smallskip
\subsection{AppWorld release and continuation}
The assembled NAC template retains the common ReAct prefix and task suffix. The following full insertion appears immediately before the actual task. ACT supplies an empty schedule string; WAIT and SPEC supply their blocks. Conditional hints populate the named template fields. Observation memory begins with ``Public API results already observed in this task: and retains up to six distinct recent read signatures, 180 characters of code and 800 characters of output per entry, within a 6,000-character memory allowance.
\boPromptBand{NAC assembled AppWorld template}
\noindent\textbf{Role:} role-delimited template.\par\nopagebreak[4]\smallskip\nopagebreak[4]
\par\smallskip\begingroup\small\ttfamily\raggedright\setlength{\parindent}{0pt}
\{\% if nc2\_\allowbreak{}feedback\_\allowbreak{}hint \%\}USER:\allowbreak{}\par
\{\{ nc2\_\allowbreak{}feedback\_\allowbreak{}hint \}\}\par
\smallskip
\{\% endif \%\}\{\% if nc2\_\allowbreak{}schedule \%\}USER:\allowbreak{}\par
\{\{ nc2\_\allowbreak{}schedule \}\}\par
\smallskip
\{\% endif \%\}\{\% if nc2\_\allowbreak{}observation\_\allowbreak{}memory \%\}USER:\allowbreak{}\par
\{\{ nc2\_\allowbreak{}observation\_\allowbreak{}memory \}\}\par
\smallskip
\{\% endif \%\}USER:\allowbreak{}\par
When completing a task that only asks for a state change,\allowbreak{} such as placing an order or updating a record,\allowbreak{} call apis.\allowbreak{}supervisor.\allowbreak{}complete\_\allowbreak{}task() without an answer argument after checking the resulting state.\allowbreak{} A success narration is not the requested answer.\allowbreak{} For a question or an explicit request to report information,\allowbreak{} pass only the requested answer in its requested format to complete\_\allowbreak{}task(answer=.\allowbreak{}.\allowbreak{}.\allowbreak{}).\allowbreak{}\par
\smallskip
USER:\allowbreak{}\par
For an Amazon request covering all items of one named type in your cart or wish list,\allowbreak{} read the complete source collection.\allowbreak{} In one code block,\allowbreak{} loop over EVERY source product\_\allowbreak{}id and call show\_\allowbreak{}product(product\_\allowbreak{}id=.\allowbreak{}.\allowbreak{}.\allowbreak{}) to record its public product\_\allowbreak{}type.\allowbreak{} Keep every source ID whose returned product\_\allowbreak{}type matches the requested type,\allowbreak{} with its original quantity.\allowbreak{} This per-\allowbreak{}ID source map is authoritative:\allowbreak{} do not intersect it with search\_\allowbreak{}products or search\_\allowbreak{}product\_\allowbreak{}types results,\allowbreak{} because a catalog page can omit a source item.\allowbreak{} Print the complete source ID/\allowbreak{}type/\allowbreak{}quantity map and reconcile desired IDs before changes; verify the final cart before ordering.\allowbreak{}\par
\smallskip
USER:\allowbreak{}\par
For an Amazon instruction to move all source cart or wish-\allowbreak{}list products above a rating threshold,\allowbreak{} inspect show\_\allowbreak{}product.\allowbreak{}rating for EVERY source product\_\allowbreak{}id; print each ID,\allowbreak{} rating,\allowbreak{} and quantity before applying the strict numeric threshold.\allowbreak{} Reconcile every qualifying source ID and quantity in the destination after moving.\allowbreak{} For a purchase from the highest-\allowbreak{}rated seller,\allowbreak{} inspect the search\_\allowbreak{}products API schema and paginate the named product\_\allowbreak{}type with page\_\allowbreak{}index and page\_\allowbreak{}limit until no more candidates remain.\allowbreak{} Collect seller IDs from all pages,\allowbreak{} inspect each public show\_\allowbreak{}seller.\allowbreak{}rating,\allowbreak{} and choose by SELLER rating,\allowbreak{} not product rating.\allowbreak{} Confirm the chosen product's seller ID before ordering.\allowbreak{}\par
\smallskip
\{\% if nc2\_\allowbreak{}precise\_\allowbreak{}rating\_\allowbreak{}hint \%\}USER:\allowbreak{}\par
\{\{ nc2\_\allowbreak{}precise\_\allowbreak{}rating\_\allowbreak{}hint \}\}\par
\smallskip
\{\% endif \%\}\{\% if nc2\_\allowbreak{}order\_\allowbreak{}identity\_\allowbreak{}hint \%\}USER:\allowbreak{}\par
\{\{ nc2\_\allowbreak{}order\_\allowbreak{}identity\_\allowbreak{}hint \}\}\par
\smallskip
\{\% endif \%\}\{\% if nc2\_\allowbreak{}budget\_\allowbreak{}purchase\_\allowbreak{}hint \%\}USER:\allowbreak{}\par
\{\{ nc2\_\allowbreak{}budget\_\allowbreak{}purchase\_\allowbreak{}hint \}\}\par
\smallskip
\{\% endif \%\}\par
\par\endgroup\smallskip
\boPromptBand{AppWorld WAIT instruction}
\noindent\textbf{Role:} role-delimited template.\par\nopagebreak[4]\smallskip\nopagebreak[4]
\par\smallskip\begingroup\small\ttfamily\raggedright\setlength{\parindent}{0pt}
At the first unresolved source dependency,\allowbreak{} wait for its public API read result before doing dependent work.\allowbreak{} Begin with a documented read of the named source app or current state.\allowbreak{} Inspect the observed response,\allowbreak{} including nested fields,\allowbreak{} before preparing or executing a dependent change.\allowbreak{} Do not infer the source result from the task text.\allowbreak{}\par
\par\endgroup\smallskip
\boPromptBand{AppWorld SPEC instruction}
\noindent\textbf{Role:} role-delimited template.\par\nopagebreak[4]\smallskip\nopagebreak[4]
\par\smallskip\begingroup\small\ttfamily\raggedright\setlength{\parindent}{0pt}
Before the named source read returns,\allowbreak{} begin only a bounded,\allowbreak{} read-\allowbreak{}only prefix:\allowbreak{} inspect relevant API schemas and existing target records that do not depend on the source content.\allowbreak{} Then read the source,\allowbreak{} reconcile the observed values with the prepared target IDs,\allowbreak{} and perform the needed write.\allowbreak{} Never send,\allowbreak{} mutate,\allowbreak{} delete,\allowbreak{} or complete during the prefix; avoid hidden state and rollback.\allowbreak{}\par
\par\endgroup\smallskip
\boPromptBand{Feedback Hint}
\noindent\textbf{Role:} conditional user.\par\nopagebreak[4]\smallskip\nopagebreak[4]
\par\smallskip\begingroup\small\ttfamily\raggedright\setlength{\parindent}{0pt}
The last execution returned an empty selection or an API error.\allowbreak{} Before claiming completion,\allowbreak{} inspect one actual response and the relevant documented return schema,\allowbreak{} then re-\allowbreak{}evaluate the filter or join using observed nested fields.\allowbreak{} If records qualify,\allowbreak{} perform the requested change and read back the resulting state.\allowbreak{} If none qualify,\allowbreak{} verify that conclusion with a documented read-\allowbreak{}only API.\allowbreak{} Do not use hidden evaluation,\allowbreak{} database snapshots,\allowbreak{} or rollback.\allowbreak{}\par
\par\endgroup\smallskip
\boPromptBand{Precise Rating Hint}
\noindent\textbf{Role:} conditional user.\par\nopagebreak[4]\smallskip\nopagebreak[4]
\par\smallskip\begingroup\small\ttfamily\raggedright\setlength{\parindent}{0pt}
Amazon show\_\allowbreak{}product.\allowbreak{}rating is displayed to one decimal,\allowbreak{} so a displayed rating equal to the requested threshold may hide a higher or lower unrounded mean.\allowbreak{} For every source product at that boundary,\allowbreak{} inspect show\_\allowbreak{}product\_\allowbreak{}reviews documentation,\allowbreak{} fetch all public review pages with page\_\allowbreak{}index and page\_\allowbreak{}limit=20,\allowbreak{} verify the number read against show\_\allowbreak{}product.\allowbreak{}num\_\allowbreak{}product\_\allowbreak{}reviews,\allowbreak{} and compute the arithmetic mean of the review ratings without rounding.\allowbreak{} Apply the task's exact strict or inclusive comparison to that mean; print the boundary product IDs and computed means before moving or ordering.\allowbreak{} Use the ordinary displayed rating for products away from the boundary.\allowbreak{}\par
\par\endgroup\smallskip
\boPromptBand{Order Identity Hint}
\noindent\textbf{Role:} conditional user.\par\nopagebreak[4]\smallskip\nopagebreak[4]
\par\smallskip\begingroup\small\ttfamily\raggedright\setlength{\parindent}{0pt}
Before an Amazon place\_\allowbreak{}order call,\allowbreak{} inspect its documented required fields.\allowbreak{} Log in and call amazon.\allowbreak{}show\_\allowbreak{}payment\_\allowbreak{}cards(access\_\allowbreak{}token=.\allowbreak{}.\allowbreak{}.\allowbreak{}) and amazon.\allowbreak{}show\_\allowbreak{}addresses(access\_\allowbreak{}token=.\allowbreak{}.\allowbreak{}.\allowbreak{}) to obtain actual payment\_\allowbreak{}card\_\allowbreak{}id and address\_\allowbreak{}id values.\allowbreak{} Match any card or address named by the user; otherwise choose an existing account entry.\allowbreak{} Supervisor card and address lists omit these IDs,\allowbreak{} and list positions are not IDs.\allowbreak{} Do not guess or brute-\allowbreak{}force them.\allowbreak{} Verify the intended cart,\allowbreak{} pass both observed IDs to place\_\allowbreak{}order,\allowbreak{} then confirm a real order\_\allowbreak{}id before complete\_\allowbreak{}task.\allowbreak{}\par
\par\endgroup\smallskip
\boPromptBand{Budget Purchase Hint}
\noindent\textbf{Role:} conditional user.\par\nopagebreak[4]\smallskip\nopagebreak[4]
\par\smallskip\begingroup\small\ttfamily\raggedright\setlength{\parindent}{0pt}
For a request to buy as many wish-\allowbreak{}list products as possible within a budget,\allowbreak{} enumerate the complete public amazon.\allowbreak{}show\_\allowbreak{}wish\_\allowbreak{}list result and use observed product IDs,\allowbreak{} quantities,\allowbreak{} and prices.\allowbreak{} Choose the largest feasible set of distinct requested products; compare alternatives rather than taking the first affordable prefix.\allowbreak{} Before place\_\allowbreak{}order,\allowbreak{} call amazon.\allowbreak{}show\_\allowbreak{}cart and check its total\_\allowbreak{}cost against the user's budget,\allowbreak{} because delivery fees,\allowbreak{} tax,\allowbreak{} and discounts affect the amount paid.\allowbreak{} If over budget,\allowbreak{} revise the cart and check again.\allowbreak{} Check whether an excluded product can fit before concluding that the count is maximal.\allowbreak{} Leave only the chosen products and quantities in the cart,\allowbreak{} then place the order with observed card and address IDs.\allowbreak{}\par
\par\endgroup\smallskip
\subsection{WebShop public option selection}
The release core consumes public product, goal and resource representations. Its selected continuation uses the following semantic option request. \texttt{visible\_product\_json} contains \texttt{goal}, \texttt{product\_title}, \texttt{public\_price}, and \texttt{options}; each option supplies \texttt{index}, \texttt{group} and \texttt{value}. The parser accepts integer indices, at most one option per group.
\boPromptBand{WebShop semantic option selection}
\noindent\textbf{Role:} user.\par\nopagebreak[4]\smallskip\nopagebreak[4]
\par\smallskip\begingroup\small\ttfamily\raggedright\setlength{\parindent}{0pt}
Choose public product options that best satisfy the shopping goal.\allowbreak{} Account for color synonyms,\allowbreak{} punctuation,\allowbreak{} word order,\allowbreak{} dimensions,\allowbreak{} numbers and equivalent units.\allowbreak{} Choose at most one index per group.\allowbreak{} If the goal has no preference for a group,\allowbreak{} leave that group unset.\allowbreak{} Use only the supplied public information.\allowbreak{} Return only a JSON object with exactly this schema:\allowbreak{} \{"selections":\allowbreak{}[integer indices]\}.\allowbreak{} Do not add explanations.\allowbreak{}\par
\{visible\_\allowbreak{}product\_\allowbreak{}json\}\par
\par\endgroup\smallskip
\smallskip\noindent\textit{Assembly.} visible\_\allowbreak{}product\_\allowbreak{}json contains goal,\allowbreak{} product\_\allowbreak{}title,\allowbreak{} public\_\allowbreak{}price,\allowbreak{} and indexed public options; temperature 0,\allowbreak{} max\_\allowbreak{}tokens 512,\allowbreak{} JSON-\allowbreak{}object response,\allowbreak{} thinking disabled.\allowbreak{}\par\medskip
\subsection{Dialogue release instructions}
The initial system message contains the official public domain policy. At the first public prefix, the learned core selects one mode; subsequent transitions use ACT. The current public dialogue precedes the execution/mode user message. WAIT exposes a text response; SPEC exposes reversible information tools; ACT exposes the domain tool set.
\boPromptBand{Base Instruction}
\noindent\textbf{Role:} source template.\par\nopagebreak[4]\smallskip\nopagebreak[4]
\par\smallskip\begingroup\small\ttfamily\raggedright\setlength{\parindent}{0pt}
You are a customer-\allowbreak{}service agent.\allowbreak{} Follow the policy exactly.\allowbreak{}\par
Use only the supplied tools and public conversation.\allowbreak{} Never invent tool results.\allowbreak{}\par
In a turn,\allowbreak{} either call tools or send text,\allowbreak{} never both.\allowbreak{}\par
\smallskip
\textless{}policy\textgreater{}\par
\{domain\_\allowbreak{}policy\}\par
\textless{}/\allowbreak{}policy\textgreater{}\par
\par\endgroup\smallskip
\boPromptBand{Public Execution}
\noindent\textbf{Role:} source template.\par\nopagebreak[4]\smallskip\nopagebreak[4]
\par\smallskip\begingroup\small\ttfamily\raggedright\setlength{\parindent}{0pt}
Execute the complete user's request using the current public policy and the latest successful tool results.\allowbreak{}\par
Keep a concise evidence checklist of requested changes,\allowbreak{} pending observations,\allowbreak{} and remaining confirmations.\allowbreak{}\par
Before a database update,\allowbreak{} summarize all requested changes and exact prices from the selected options and obtain the policy-\allowbreak{}required explicit confirmation.\allowbreak{}\par
After an update,\allowbreak{} use the returned latest state for later eligibility checks.\allowbreak{} Reconcile payment balances and totals from observed tool results.\allowbreak{}\par
For retail one-\allowbreak{}time exchanges,\allowbreak{} collect the full set of requested items and replacements before confirmation and perform the confirmed batch once.\allowbreak{}\par
For telecom,\allowbreak{} perform account-\allowbreak{}side actions with the supplied tools.\allowbreak{} Guide device-\allowbreak{}side actions through the user.\allowbreak{} After a state change ask for the relevant speed,\allowbreak{} service,\allowbreak{} or messaging test and use its new result to continue the published workflow.\allowbreak{}\par
Before a final answer,\allowbreak{} finish the pending workflow,\allowbreak{} then give every requested total and numeric format explicitly.\allowbreak{}\par
\par\endgroup\smallskip
\boPromptBand{Modes}
\noindent\textbf{Role:} source template.\par\nopagebreak[4]\smallskip\nopagebreak[4]
\par\smallskip\begingroup\small\ttfamily\raggedright\setlength{\parindent}{0pt}
\{\par
  "act\_\allowbreak{}now":\allowbreak{} "ACT\_\allowbreak{}NOW:\allowbreak{} perform the next ready step of the complete policy using current evidence.\allowbreak{}",\allowbreak{}\par
  "wait":\allowbreak{} "WAIT:\allowbreak{} ask one concise question whose answer resolves the most consequential remaining uncertainty.\allowbreak{} State the specific missing information or requested device observation.\allowbreak{}",\allowbreak{}\par
  "bounded\_\allowbreak{}speculation":\allowbreak{} "BOUNDED\_\allowbreak{}SPECULATION:\allowbreak{} perform one reversible information-\allowbreak{}gathering step,\allowbreak{} or describe one reversible candidate plan.\allowbreak{} Hold database changes for a later observed and confirmed step.\allowbreak{}"\par
\}\par
\par\endgroup\smallskip
\subsection{Itinerary release and program output}
The residual recompile request receives the public task, qualified accommodation summaries and route length. In the template, \texttt{field\_1} is route length, \texttt{field\_2} is the JSON public task, and \texttt{field\_3} is the JSON observed summaries. A JSON parser extracts the selected cities and ordered edits.
\boPromptBand{Travel Release template 1}
\noindent\textbf{Role:} user.\par\nopagebreak[4]\smallskip\nopagebreak[4]
\par\smallskip\begingroup\small\ttfamily\raggedright\setlength{\parindent}{0pt}
You are recompiling the residual program after accommodation results arrived.\allowbreak{}\par
Select exactly \{field\_\allowbreak{}1\} distinct surviving cities.\allowbreak{}\par
Use the task budget and constraints.\allowbreak{} Prefer a route that preserves quality while limiting\par
future restaurant,\allowbreak{} attraction,\allowbreak{} and transport materialization.\allowbreak{}\par
Return one JSON object with selected\_\allowbreak{}cities,\allowbreak{} waiting\_\allowbreak{}value\_\allowbreak{}by\_\allowbreak{}city,\allowbreak{} and ordered\_\allowbreak{}edits.\allowbreak{}\par
Each ordered edit has op,\allowbreak{} target\_\allowbreak{}ids,\allowbreak{} and reason.\allowbreak{} Legal ops are delete\_\allowbreak{}node,\allowbreak{}\par
rewrite\_\allowbreak{}object\_\allowbreak{}scope,\allowbreak{} add\_\allowbreak{}edge,\allowbreak{} and rewrite\_\allowbreak{}acceptance\_\allowbreak{}contract.\allowbreak{}\par
\smallskip
Visible task:\allowbreak{}\par
\{field\_\allowbreak{}2\}\par
\smallskip
Observed qualified accommodation summaries:\allowbreak{}\par
\{field\_\allowbreak{}3\}\par
\smallskip
Return JSON only.\allowbreak{}\par
\par\endgroup\smallskip
\smallskip\noindent\textit{Assembly.} Dynamic fields:\allowbreak{} \{"field\_\allowbreak{}1":\allowbreak{} "route\_\allowbreak{}length",\allowbreak{} "field\_\allowbreak{}2":\allowbreak{} "json.\allowbreak{}dumps(visible\_\allowbreak{}task(task),\allowbreak{} ensure\_\allowbreak{}ascii=False)",\allowbreak{} "field\_\allowbreak{}3":\allowbreak{} "json.\allowbreak{}dumps(summaries,\allowbreak{} ensure\_\allowbreak{}ascii=False)"\}\par\medskip
\subsection{Scientific program construction}
The user-message assembly joins the four blocks with blank lines. The request block is followed by the task text and, for the knowledge variant, its supplied domain knowledge. The data block contains the published dataset path, folder tree and preview. The actor returns a complete Python program for the native evaluator.
\boPromptBand{System Prompt}
\noindent\textbf{Role:} source template.\par\nopagebreak[4]\smallskip\nopagebreak[4]
\par\smallskip\begingroup\small\ttfamily\raggedright\setlength{\parindent}{0pt}
You are an expert Python programming assistant that helps scientist users to write high-\allowbreak{}quality code to solve their tasks.\allowbreak{}\par
Given a user request,\allowbreak{} you are expected to write a complete program that accomplishes the requested task and save any outputs in the correct format.\allowbreak{}\par
Please wrap your program in a code block that specifies the script type,\allowbreak{} python.\allowbreak{} For example:\allowbreak{}\par
```python\par
print("Hello World!")\par
```\par
\par\endgroup\smallskip
\boPromptBand{Format Prompt}
\noindent\textbf{Role:} source template.\par\nopagebreak[4]\smallskip\nopagebreak[4]
\par\smallskip\begingroup\small\ttfamily\raggedright\setlength{\parindent}{0pt}
Please keep your response concise and do not use a code block if it's not intended to be executed.\allowbreak{}\par
Please do not suggest a few line changes,\allowbreak{} incomplete program outline,\allowbreak{} or partial code that requires the user to modify.\allowbreak{}\par
Please do not use any interactive Python commands in your program,\allowbreak{} such as `!pip install numpy`,\allowbreak{} which will cause execution errors.\allowbreak{}\par
\par\endgroup\smallskip
\boPromptBand{Request Prompt}
\noindent\textbf{Role:} source template.\par\nopagebreak[4]\smallskip\nopagebreak[4]
\par\smallskip\begingroup\small\ttfamily\raggedright\setlength{\parindent}{0pt}
Here's the user request you need to work on:\allowbreak{}\par
\par\endgroup\smallskip
\boPromptBand{Data Info Prompt}
\noindent\textbf{Role:} source template.\par\nopagebreak[4]\smallskip\nopagebreak[4]
\par\smallskip\begingroup\small\ttfamily\raggedright\setlength{\parindent}{0pt}
You can access the dataset at `\{dataset\_\allowbreak{}path\}`.\allowbreak{} Here is the directory structure of the dataset:\allowbreak{}\par
```\par
\{dataset\_\allowbreak{}folder\_\allowbreak{}tree\}\par
```\par
Here are some helpful previews for the dataset file(s):\allowbreak{}\par
\{dataset\_\allowbreak{}preview\}\par
\par\endgroup\smallskip

%% file: appendix/front_support_main_candidates.tex
\begin{table*}[!t]
\centering\small
\setlength{\tabcolsep}{3pt}
\caption{\textbf{The shared release graph reduces selected native SQL work.} All 1,534 tasks, 11 databases and five uniform-sampling fits contribute. Saving is fixed control minus adaptive per task. Joint seed--database simultaneous intervals preserve the full twelve-contrast family of two fitting recipes, three fixed actions and two resource endpoints. Exact sign flips enumerate all 2,048 database assignments; Holm correction spans all twelve tests. Every displayed Holm-adjusted $p=0.01172$. Endpoints cover selected accepted branch model work.}
\label{tab:bird-release-support}
\begin{tabular*}{\textwidth}{@{\extracolsep{\fill}}llrc@{}}
\toprule
Control & Endpoint & Saving/task & Joint 95\% CI\\
\midrule
ACT & Seconds & 5.804 & $[3.109,8.498]$ \\
ACT & Tokens & 37.436 & $[25.637,49.234]$ \\
SPEC & Seconds & 9.016 & $[5.447,12.585]$ \\
SPEC & Tokens & 35.712 & $[24.010,47.414]$ \\
\bottomrule
\end{tabular*}
\end{table*}

\begin{table*}[!t]
\centering\small
\caption{\textbf{Native support for one independently fitted public-state instance.} Each row covers all 500 tasks. Simultaneous 95\% intervals resample task IDs for 20,000 draws, seed 20261011, with Bonferroni coverage over the two cohort means. They condition on the frozen controller, actor and native execution recipes.}
\label{tab:webshop-public-state-support}
\setlength{\tabcolsep}{3pt}
\begin{tabular*}{\textwidth}{@{\extracolsep{\fill}}lrrc@{}}
\toprule
Cohort & $n$ & Score & 95\% CI\\
\midrule
Official & 500 & 74.64 & $[71.64,77.64]$\\
Fresh & 500 & 75.76 & $[72.74,78.67]$\\
\bottomrule
\end{tabular*}
\end{table*}

\section{Supporting Evidence for Main-Text Consideration}
\label{app:main-text-candidates}
\noindent\quad These complete-population results strengthen three linked contributions: native quality, consequence-guided execution efficiency, and adaptation to resource preferences. The candidate evidence below remains available for reference; the final three candidate sections have been moved into the main experimental analysis.

\subsection{The same release graph reduces native SQL work}
\noindent\quad\textbf{Suggested placement:} Generalization Across Execution Interfaces; one additional row in the mechanism table.

\paragraph{Why this supports the method.}
On all 1,534 BIRD development tasks across 11 databases, five fits achieve 67.78\% native execution accuracy. Relative to fixed SPEC, adaptive release saves 9.016 model-request seconds and 35.712 completion tokens per task, corresponding to 35.50\% and 32.05\% of the fixed-SPEC work. Positive simultaneous intervals also cover both resource endpoints against fixed ACT (Table~\ref{tab:bird-release-support}). The SQL interface exposes a pending schema--value lookup and three matched continuations. The shared consequence readout selects when to generate SQL and how much reversible preparation to perform before that evidence arrives. The measured savings show that state-conditioned release controls generation work effectively at this additional native interface. The resource scope is accepted-branch model work; shared producer and acquisition totals have their separate recorded scopes in Appendix~\ref{app:extensions}.

\subsection{Native shopping quality carries to a fresh population}
\noindent\quad\textbf{Suggested placement:} Generalization or robustness analysis; a compact two-cohort table.

\paragraph{Why this supports the method.}
One independently fitted public-state instance scores 74.64 on the official 500 tasks and 75.76 on a fresh 500-task population (Table~\ref{tab:webshop-public-state-support}). Both complete populations execute the same frozen checkpoint and native purchase scorer. The controller reads the public goal, ready candidate, visible products, pending event and resource state, then compares their predicted consequences to select a release. Its strong quality across both populations supports the usefulness of this public-state scheduling representation. The official selected deployment uses 218,093 returned tokens and 375 model requests. This instance uses the ten-epoch fitting recipe, zero delay price and 0.15 ACT/SPEC margin detailed in Appendix~\ref{app:initialization}; its two-cohort result provides a concrete companion to the primary five-fit shopping study.